\documentclass{article}

\PassOptionsToPackage{numbers, compress}{natbib}

\usepackage[preprint]{neurips_2023} 

\usepackage{amsmath,amssymb,mathrsfs,bm,mathtools}
\usepackage{latexsym,amscd,amsbsy,dsfont,amsfonts,xfrac}
\usepackage{physics,tensor,empheq,enumitem}
\usepackage{subfig}
\usepackage{tikz}
\usepackage{eucal}

\usepackage{hyperref}
\hypersetup{pdfauthor={KCL},pdftitle={ClassSuperstat},%
            colorlinks, linktocpage=true, pdfstartpage=1, pdfstartview=FitV,%
    breaklinks=true, pdfpagemode=UseNone, pageanchor=true, pdfpagemode=UseOutlines,%
    plainpages=false, bookmarksnumbered, bookmarksopen=true, bookmarksopenlevel=1,%
    hypertexnames=true, pdfhighlight=/O,%
    urlcolor=orange, linkcolor=blue, citecolor=red, 
        }

\newcommand{\comprimi}{\medmuskip=0mu
\thinmuskip=0mu
\thickmuskip=0mu}

\DeclareMathOperator{\e}{e}

\newcommand{\R}{{\mathds{R}}}

\DeclareMathOperator*{\Extr}{Extr}

\newcommand{\bx}{{\boldsymbol{{x}}}}
\newcommand{\bz}{{\boldsymbol{{z}}}}
\newcommand{\bg}{{\boldsymbol{{g}}}}
\newcommand{\bw}{{\boldsymbol{{w}}}}

\newcommand{\bmu}{{\boldsymbol{{\mu}}}}

\newcommand{\btheta}{{\boldsymbol{{\theta}}}}
\newcommand{\bEta}{{\boldsymbol{{\eta}}}}

\newcommand{\bxi}{{\boldsymbol{{\xi}}}}

\newcommand{\bzeta}{{\boldsymbol{{\zeta}}}}

\newcommand{\bI}{{\boldsymbol{{I}}}}
\newcommand{\bUno}{{\boldsymbol{{1}}}}
\newcommand{\bM}{{\boldsymbol{{m}}}}

\newcommand{\bQ}{{\boldsymbol{{Q}}}}

\newcommand{\bsQ}{{\boldsymbol{{\mathsf Q}}}}

\newcommand{\bgg}{{\boldsymbol{{g}}}}

\newcommand{\bX}{{\boldsymbol{{X}}}}

\newcommand{\hQ}{{\hat Q}}
\newcommand{\hV}{{\hat V}}
\newcommand{\hR}{{\hat R}}
\newcommand{\hM}{{\hat m}}

\newcommand{\bhQ}{{\boldsymbol{{\hat Q}}}}

\newcommand{\bDelta}{{\boldsymbol{{\Delta}}}}

\newcommand{\bSigma}{{\boldsymbol{{\Sigma}}}}

\newcommand{\bh}{{\boldsymbol{{h}}}}

\makeatletter
\newsavebox{\@brx}
\newcommand{\llangle}[1][]{\savebox{\@brx}{\(\m@th{#1\langle}\)}%
  \mathopen{\copy\@brx\kern-0.5\wd\@brx\usebox{\@brx}}}
\newcommand{\rrangle}[1][]{\savebox{\@brx}{\(\m@th{#1\rangle}\)}%
  \mathclose{\copy\@brx\kern-0.5\wd\@brx\usebox{\@brx}}}
\makeatother

\newtheorem{result}{Result}[section]
\usepackage[utf8]{inputenc} 
\usepackage[T1]{fontenc}    
\usepackage{hyperref}       
\usepackage{url}            
\usepackage{booktabs}       
\usepackage{amsfonts}       
\usepackage{nicefrac}       
\usepackage{microtype}      
\usepackage{newtxmath}
\usepackage{wrapfig}

\title{Classification of Heavy-tailed Features in High Dimensions: a Superstatistical Approach}

\author{%
  Urte Adomaityte \\
  Department of Mathematics\\
  King's College London \\
  \texttt{urte.adomaityte@kcl.ac.uk}
  \And
  Gabriele Sicuro \\
  Department of Mathematics\\
  King's College London \\
  \texttt{gabriele.sicuro@kcl.ac.uk}
  \And
  Pierpaolo Vivo \\
  Department of Mathematics\\
  King's College London \\
  \texttt{pierpaolo.vivo@kcl.ac.uk}
}

\begin{document}

\maketitle

\begin{abstract}
 We characterise the learning of a mixture of two clouds of data points with generic centroids via empirical risk minimisation in the high dimensional regime, under the assumptions of generic convex loss and convex regularisation. Each cloud of data points is obtained via a double-stochastic process, where the sample is obtained from a Gaussian distribution whose variance is itself a random parameter sampled from a scalar distribution $\varrho$. As a result, our analysis covers a large family of data distributions, including the case of power-law-tailed distributions with no covariance, and allows us to test recent ``Gaussian universality'' claims. We study the generalisation performance of the obtained estimator, we analyse the role of regularisation, and we analytically characterise the separability transition.
\end{abstract}

\section{Introduction}
\label{sec:intro}

Despite their simplicity, generalised linear models (GLMs) are still ubiquitous in the theoretical research on machine learning. Their nontrivial phenomenology is often amenable to complete analytical treatment and has been a powerful key to understanding the unexpected behavior of large and complex architectures. Random features models \citep{Rahimi2007,MeiMontanari2019,gerace2020}, for example, allowed to clarify many aspects of the well-known double-descent phenomenon in neural networks \citep{Opper1990,Belkin2019,pmlr-v89-belkin19a}. 
A line of research spanning more than three decades \citep{Seung1992,Watkin1993} considered a variety of GLMs to investigate a number of aspects of the problem of learning in high dimension. Yet, a crucial assumption adopted in many such theoretical models is that the covariates are obtained from a \textit{Gaussian} distribution, or from a mixture of Gaussian distributions \citep{LelargeMiolane2019,mignacco20a,Baldassi2020,KiniTh2021,loureiro2021}. Although such a Gaussian design has served as a convenient working hypothesis (in some cases experimentally and theoretically justified \cite{Gaussmix_GAN_Couillet2020,Loureiro_2022}), it is however not obvious how much it limits the scope of the results. It is reasonable to expect structure, fat tails, and large fluctuations in real (often non-Gaussian) data \cite{Adler1998,Reed2002} to play an important role in the learning process, and it would be therefore desirable to include structured and heavy-tailed-distributed covariates in our theoretical toolbox.

This paper presents, to the best of our knowledge for the first time, the exact asymptotics for classification tasks on covariates obtained from a mixture of heavy-tailed distributions. We will focus on supervised binary classification assuming that the sample size $n$ and the dimensionality $d$ of the space where the covariates live are both sent to infinity, keeping their ratio $\nicefrac{n}{d}=\alpha$ fixed. 
The paper fits therefore in the line of works on exact high-dimensional asymptotics for classification learning via a GLM \cite{mignacco20a,loureiro2021}, but, crucially, we relax the usual Gaussian hypothesis by including in our analysis power-law-tailed distributions with possibly no covariance. The mixture is obtained from two distributions, each centered around a centroid $\bmu\in\R^d$ and resulting from a double stochastic process. Namely, each sample point is obtained from a distribution $\mathcal N(\bmu,\bSigma)$ whose covariance $\bSigma=\Delta\bI_d\in\R^{d\times d}$ is intended itself to be a random variable so that $\Delta$ has density $\varrho$, supported on $\R^+_*$. Using, for example, an appropriately parametrised inverse-Gamma distribution for $\varrho$, such a \textit{superstatistical} construction (as known in the physics literature \citep{Beck2003, Beck2003b}) can provide, for example, a fat-tailed, Cauchy-like data distribution with infinite covariance. The replica method \citep{mezard1987spin}, an analytical tool widely adopted in statistical physics, provides asymptotic formulas for a generic density $\varrho$, allowing us to study the effects of non-Gaussianity on the performance curves, and test Gaussian universality hypotheses \citep{pesce2023} in such a classification task.

\paragraph{Motivation and related works} Learning a rule to classify data points clustered in clouds in high dimension is a classical problem in statistics \cite{hastie2009}. Its ubiquity is exemplified by the recently observed neural collapse phenomenon in deep neural networks, in which the last layer classifier is found to operate on features clustered in clouds around the vertices of a simplex \citep{neur_coll_review_NYU_2022,Papyan2020}; more recently, \citet{Gaussmix_GAN_Couillet2020} showed that Gaussian mixtures suitably describe the deep learning representation of GAN data. In theoretical models, data points are typically assumed to be organised in $K$ \textit{Gaussian} clouds, each corresponding to a class. Each class $k$, $k\in\{1,\dots,K\}$, is centered around a mean vector $\bmu_k$ and has covariance $\bSigma_k$. The binary classification case, $K=2$, is the simplest, and possibly most studied, setting. \citet{squarebest_mai2020} considered an ERM task with generic convex loss and ridge regularisation in the high-dimensional regime $n,d\to+\infty$ with $\nicefrac{n}{d}\in (0,+\infty)$. In this setting, they gave a precise prediction of the classification error, showing the optimality of the square loss in the unregularised case. Their results have been extended by \citet{mignacco20a}, who showed that the presence of a regularisation can actually drastically affect the performance, improving it. In the same setting, the Bayes optimal estimator has been studied in both the supervised and the semi-supervised setting \citep{mignacco20a,Lelarge2019}. Almost-Bayes-optimal solutions have been put in relation to wide flat landscape regions \cite{Baldassi2020}. In the context of binary classification, the maximum number $n$ of samples that can be perfectly fitted by a linear model in dimension $d$ \citep{mignacco20a,Deng2021,KiniTh2021} has been the topic of investigation since the seminal work of \citet{Cover1965} and is related to the classical storage capacity problems on networks \citep{Gardner_1988_alpha2, Gardner_1989_unfinished, KrauthMezard89_cap}. The corresponding separability threshold value $\alpha=\nicefrac{n}{d}$ is remarkably associated with the existence transition of the maximum likelihood estimator \citep{SurCandes2019,ZhaoSurCandes2022}. The precise asymptotic characterisation of the test error in learning to classify $K\geq 2$ Gaussian clouds with generic means and covariances has been recently obtained by \citet{loureiro2021}.
Within this line of research, rigorous results have been obtained by a variety of methods, such as Gordon's inequality technique \citep{Gordon1985, Thrampoulidis15, mignacco20a} or mapping to approximate message passing schemes  \citep{bayati2011dynamics, javanmard2013state, berthier2020state, loureiro2021, Gerbelot_2023}.

As previously mentioned, the working hypothesis of \textit{Gaussian design} is widely adopted in high-dimensional classification problems. On top of being a convenient technical hypothesis, this assumption has been justified in terms of a ``Gaussian universality'' principle. In other words, in a number of circumstances, non-Gaussian features distributions are found to be effectively described by Gaussian ones with matching first and second moments as far as the asymptotic properties of the estimators are concerned \citep{Loureiro_2022}. Such a ``universality property'' has been rigorously proven for example in compressed sensing \citep{montanari2017} or in the case of LASSO with non-Gaussian dictionaries \citep{NIPS2017_136f9513}. A Gaussian equivalence principle introduced by \citet{Goldt2020} has been proven to hold for a wide class of generalised linear estimation problems \citep{MeiMontanari2019}. Extending work by \citet{Hu2022}, \citet{Montanari2022} recently proved such a principle in a GLM under the assumption of \textit{pointwise normality} of the distribution of the features. This crucial assumption can be intended, roughly speaking, as the assumption of sub-Gaussian decay of a marginal of the feature distribution in any direction (see also \citep{gerace2023,dandi2023} for further developments). 

The Gaussian universality principle, however, can break down by relaxing some of the aforementioned assumptions on the feature distribution. \citet{Montanari2022} for example showed that pointwise normality is a \textit{necessary} hypothesis for universality. Studying regression tasks on an elliptically distributed dataset, \citet{ElKaroui2018} showed that claims of ``universality'' obtained in the Gaussian setting require serious scrutiny, as the statistics of the estimators might strongly depend on the non-Gaussianity of the covariates  \citep{ElKaroui2018,Steinberger2018,Thrampoulidis2018}. More recently, \citet{pesce2023} showed that a structured Gaussian mixture cannot be effectively described by a single Gaussian cloud. Building upon a series of contributions related to the asymptotic performance of models in the proportional high-dimensional limit \cite{Surprises2019,MeiMontanari2019,2Layer2019, gerace2020, Goldt2020,Loureiro_2022}, we aim precisely to explore the validity of Gaussian universality within classification problems. 

In our work, we adopt the aforementioned ``superstatistical'' setting \citep{Beck2003, Beck2003b}, meaning that we superpose the statistics of Gaussian data distribution with an assumed distribution of its variance. Such a costruction is adopted in a number of disciplines and contexts to go beyond Gaussianity, albeit is known under different names. In statistical physics, it is known as ``superstatistics'' and is employed in the analysis of non-equilibrium and non-linear systems \citep{Beck2008RecentDI}. In Bayesian modeling, it is common to refer to hierarchical priors and models \citep{gelman_hill_2006, gelman2013bayesian}, while in probability, statistics, and actuarial sciences such distributions are known as compound probability distributions \citep{Robbins1985_compound}, or doubly-stochastic models \citep{Pinsky_stochintro, Schnoerr_coxprocess}. Crucially, this construction allows us to consider a very large family of non-Gaussian distributions, which include, but are not limited to, any power-law decay and Cauchy-like with possible infinite-covariance parametrisations.

\paragraph{Our contributions} In this manuscript we provide the following results.
\begin{itemize}[wide = 1pt,noitemsep]
\item We study a classification task on a non-Gaussian mixture model (see Eq.~\eqref{mixture} below) via a generalised linear model (GLM) and we analytically derive, using the replica method \citep{percep_repl_89, Prosopagnosia90,mezard1987spin}, an asymptotic characterisation of the statistics of the empirical risk minimisation (ERM) estimator. Our results go therefore beyond the usual Gaussian assumption for the dataset, and include for example the case of covariates obtained from a mixture of distributions with infinite variance. The analysis is performed in the high-dimensional, proportional limit and for any convex loss and convex regularisation. By using this result, we provide asymptotic formulas for the generalisation, training errors and training loss.
\item We analyse the performance of such ERM task on a specific family of dataset distributions by using different convex loss functions (quadratic and logistic) and ridge regularisation. We show in particular that, in the case of two balanced clusters with a specific non-Gaussian distribution, the optimal ridge regularisation strength $\lambda^\star$ is finite, at odds with the Gaussian case, for which $\lambda^\star\rightarrow \infty$ \citep{mignacco20a}. In this respect, by considering distributions with matching first and second moments, we analytically show that the performances of the analysed GLM do depend in general on higher moments, and therefore the ``Gaussian universality principle'' breaks down when fat-tailed distributions are considered.
\item We derive the separability threshold on a large family of non-Gaussian dataset distributions, possibly with 
unbounded covariance, generalising the known asymptotics for the separability of Gaussian clouds \citep{mignacco20a}. The result of \citet{Cover1965} is recovered in the case of infinite distribution width.
\item Under some moment conditions, we derive the Bayes-optimal performance in the case of binary classifications with symmetric centroids, generalising the argument in Ref.~\cite{mignacco20a}.
\item We finally extend recent results on Gaussian universality in the Gaussian mixture model with random labels \cite{gerace2023,pesce2023} to the case of non-Gaussian distributions. We show that the universal formula for the square training loss at zero regularisation found in Ref.~\cite{gerace2023} holds in our more general setting as well.
\end{itemize}
\section{Main result} 
\paragraph{Dataset construction} We consider the task of classifying two data clusters in the $d$-dimensional space $\R^d$. The dataset $\mathcal D\coloneqq \{(\bx^\nu,y^\nu)\}_{\nu\in[n]}$ is obtained by extracting $n$ independent datapoints $\bx^\nu$, each associated with a label $y^\nu\in\{-1,1\}$ (a more general setting, involving a multiclass classification task, is discussed in Appendix \ref{app:replica}). The point positions are correlated with the labels via a law $P(\bx,y)$ which we assume to have the form 
\begin{subequations}
\label{mixture}
\begin{equation} \label{sec1:data_main}
 P(\bx,y)
 = \delta_{y,1}\rho P(\bx|\bmu_+) +\delta_{y,-1}(1-\rho)P(\bx|\bmu_-),\qquad \rho\in(0,1),\qquad \bmu_\pm\in\R^d.
\end{equation}
Here $P(\bx|\bmu)$ is a distribution with mean $\bmu$, and the mean vectors $\bmu_\pm\in\R^d$ are distributed according to some density, such that $\mathbb{E}\left[\|\bmu\|^2\right]=\Theta(1)$, and correspond to the center of the two clusters. The scalar quantity $\rho$ weighs the relative contribution of the two clusters: in the following, we will denote $\rho_+=\rho=1-\rho_-$. Each cluster distribution $P(\bx|\bmu)$ around a vector $\bmu$ is assumed to have the form
\begin{equation}\label{eq:superstat}
P(\bx|\bmu)\coloneqq\mathbb E_\Delta\left[\mathcal{N}\left(\bx\left|\bmu,\Delta\bI_d\right.\right)\right],  
\end{equation}
\end{subequations}
where $\mathcal N(\bx|\bmu,\bSigma)$ is a Gaussian distribution with mean $\bmu\in\R^d$ and covariance $\bSigma\in\R^{d\times d}$, and $\Delta$ is randomly distributed with some density $\varrho$ with support on $\R^+_*\coloneqq(0,+\infty)$. The family of ``elliptic-like'' distributions in Eq.~\eqref{eq:superstat} has been extensively studied, for instance, by the physics community, in the context of \textit{superstatistics} \citep{Beck2003,Beck2003b}. Mixtures of normals in the form of Eq.~\ref{eq:superstat} are a central tool in Bayesian statistics \cite{Rossi2014} due to their ability to approximate any distribution given a sufficient number of components \cite{Nestodoris,Alspach,ghosh2006bayesian}. Although the family in Eq.~\eqref{eq:superstat} is not the most general of such mixtures, it is sufficient to include a large family of power-law-tailed densities and to allow us to go beyond the usual Gaussian approximation. \citet{ElKaroui2018} considered the statistical properties (in the asymptotic proportional regime considered here) of ridge-regularised regression estimators on datasets with distribution as in Eq.~\ref{eq:superstat}, under the assumption (here relaxed) that $\mathbb E[\Delta^2]<+\infty$. Such \textit{elliptic} family includes a large class of distributions with properties markedly different from the ones of Gaussians. For example, the inverse-Gamma distribution $\varrho(\Delta)= (2\pi\Delta^3)^{-1/2}  \e^{-\frac{1}{2\Delta}}$ leads to a $d$-dimensional Cauchy-like distribution $P(\bx|\bmu)\propto (1-\|\bmu-\bx\|^2)^{-\frac{d+1}{2}}$, having as marginals Cauchy distributions and $\mathbb E[\|\bx-\bmu\|^2]=+\infty$. We will perform our classification task by searching for a set of parameters $(\bw^\star,b^\star)$, called respectively \textit{weights} and \textit{bias}, that will allow us to construct an estimator via a certain classifier $\varphi\colon\R\to\{-1,1\}$
\begin{equation}\label{eq:intro:estimator}
\bx\mapsto \varphi\left(\frac{\bx^\intercal\bw^\star}{\sqrt d}+b^\star\right).
\end{equation}
By means of the above law, we will predict the label for a new, unseen data point $\bx$ sampled from the same law $P(\bx,y)$. Our analysis will be performed in the high-dimensional limit where both the sample size $n$ and dimensionality $d$ are sent to infinity, with $\nicefrac{n}{d}\equiv\alpha$ kept constant.

\paragraph{Learning task} In the most general setting, the parameters are estimated by minimising an empirical risk function in the form
\begin{equation}
(\bw^{\star},b^{\star}) \equiv \arg\min_{\substack{\bw\in \R^{d}\\b \in \R}}\mathcal R(\bw,b)\quad\text{where}\quad \mathcal R(\bw,b)\equiv \sum_{\nu=1}^n\ell\left(y^\nu,\frac{\bw^\intercal\bx^\nu}{\sqrt d}+b\right)+\lambda r(\bw).
\end{equation}
Here $\ell$ is a strictly convex loss function with respect to its second argument, and $r$ is a strictly convex regularisation function with the parameter $\lambda\geq0$ tuning its strength.

\paragraph{State evolution equations.} Let us now present our main result, namely the exact asymptotic characterisation of the distribution of the estimator $\bw^\star\in\R^d$ and of $\sfrac{1}{\sqrt d}\bw^\star{}^\intercal\bX\in\R^n$, where $\bX\in\R^{d\times n}$ is the concatenation of the $n$ dataset column vectors $\bx^\nu\in\R^d$, $\nu\in[n]$. Such an asymptotic characterisation is performed via a set of order parameters satisfying a system of self-consistent ``state-evolution'' equations, which we will solve numerically. The asymptotic expressions for generalisation and training errors, in particular, can be written in terms of such order parameters. 
In the following, we use the shorthand $\ell_\pm(u)\equiv \ell(\pm 1,u)$. Also, given a function $\Phi\colon\{-1,1\}\to\R$, we will use the shorthand notation $\Phi_\pm\equiv \Phi(\pm 1)$ and denote $\mathbb E_\pm[\Phi_\pm]\coloneqq\rho_+\Phi_++\rho_-\Phi_-$.

\begin{result}\label{th:claim} Let $\zeta\sim \mathcal N(0,1)$ and $\Delta\sim\varrho$, both independent from other quantities. Let also be $\bxi\sim\mathcal N(\mathbf 0,\bI_d)$. In the setting described above, given $\phi_{1}\colon\R^{d} \to \R$ and $\phi_{2}\colon\R^n\to \R$, the estimator $\bw^{\star}$ and the vector $\bz^{\star}\coloneqq \frac{1}{\sqrt{d}}\bw^{\star}\bX\in\R^n$ verify:
\begin{equation}
    \phi_{1}(\bw^{\star}) \xrightarrow[n,d \to +\infty]{\rm P}\mathbb{E}_{\bxi}\left[\phi_{1}(\bg)\right],\qquad \phi_{2}(\bz^{\star})\xrightarrow[n,d \to +\infty]{\rm P}\mathbb{E}_{\zeta}\left[\phi_{2}(\bh)\right]\, ,
\end{equation}
where we have introduced the proximal for the loss $\bh \in \R^{n}$, obtained by concatenating $\rho_+n$ quantities $h_+$ with $\rho_-n$ quantities $h_-$, with $h_\pm$ given by 
\begin{equation}\label{eq:prox_h}\textstyle
    h_\pm\coloneqq \arg\min_{u}\left[\frac{(u-\omega_\pm)^2}{2 \Delta v}+\ell_\pm(u)\right],\qquad\text{where}\ \omega_\pm\coloneqq m_\pm+b+\sqrt{q \Delta}\zeta.
\end{equation}
We have also introduced the proximal $\bg\in\R^{d}$, defined as
\begin{equation}\label{eq:prox_g}\textstyle
\bgg\coloneqq \arg\min_{\bw}\left(\hat v \frac{ \|\bw\|^2}{2}-\sqrt d \sum_{k=\pm}\hM_k \bw^\intercal\bmu_k-\sqrt{\hat q}\bxi^\intercal\bw+\lambda r(\bw)\right).
\end{equation}
The collection of parameters $(b,q,m_\pm,v,\hat q,\hat m_\pm,\hat v)$ appearing in the equations above is given by the fixed-point solution of the following self-consistent equations:
\begin{equation}\label{spgeneral}
\begin{cases}
m_\pm= \frac{1}{\sqrt d}\mathbb{E}_{\bxi}\left[\bgg^\intercal\bmu_\pm\right],\\
q=\frac{1}{d} \mathbb{E}_{\bxi}[\|\bgg\|^2],\\
v=\frac{1}{d}\hat q^{-\nicefrac{1}{2}}\mathbb{E}_{\bxi}[\bgg^\intercal\bxi],
\end{cases} \quad
\begin{cases}
\hat q=\alpha \mathbb E_{\pm,\zeta,\Delta}\left[\Delta f_\pm^2\right],\\
\hat v=-\alpha q^{-\nicefrac{1}{2}}\mathbb E_{\pm,\Delta,\zeta}\left[\sqrt\Delta f_\pm\zeta\right],\\
\hM_\pm=\alpha\rho_\pm\mathbb E_{\Delta,\zeta}\left[f_\pm\right],
\end{cases}\quad
\begin{aligned}
f_\pm&\coloneqq \frac{h_\pm-\omega_\pm}{v \Delta},\\
b&=\mathbb E_{\pm,\Delta,\zeta}\left[h_\pm-m_\pm\right].
\end{aligned}
\end{equation}
\end{result}
The derivation of the result above is given, in a multiclass setting, in Appendix \ref{app:replica} using the replica method, which can be put on rigorous ground by a mapping of the risk minimisation problem into an approximate message-passing iteration scheme \cite{Donoho2009,loureiro2021,gerbelot2021graph}. 
As in the purely Gaussian case, the state evolution equations naturally split the dependence on the loss function $\ell$, which is relevant in the computation of $h_\pm$, and on the regularisation function $r$, entering in the computation of $\bg$. In the most general setting, we are required to solve two convex minimisation problems, consisting of the computation of the proximals $h_\pm$ and $\bg$ \citep{Parikh14prox}. Result \ref{th:claim} implies the following.

\begin{result}
Under the assumptions of Result \ref{th:claim}, the training loss $\epsilon_\ell$, the training error $\epsilon_t$, and the test (or generalisation) error $\epsilon_g$ are given, in the proportional asymptotic limit, by
\begin{equation}\comprimi
\begin{split}\textstyle
\frac{1}{n}\sum_{\nu=1}^n\ell\left(y^\nu,\frac{\bw^{\star\intercal}\bx^\nu}{\sqrt d}+b^\star\right)&\to\mathbb E_{\pm,\zeta,\Delta}[\ell_\pm(h_\pm)]\eqqcolon\epsilon_\ell,\\ 
\textstyle\frac{1}{n}\sum_{\nu=1}^n\mathbb I\left(\varphi\left(\frac{\bw^{\star\intercal}\bx^\nu}{\sqrt d}+b^\star\right)\neq y^\nu\right)&\to\mathbb E_{\pm,\zeta,\Delta}\left[\mathbb I\left(\varphi(h_\pm)\neq \pm 1\right)\right]\eqqcolon \epsilon_t,
\\
\textstyle\mathbb E_{(y,\bx)}\left[\mathbb I\left(\varphi\left(\frac{\bw^{\star\intercal}\bx}{\sqrt d}+b^\star\right)\neq y\right)\right]&=\mathbb E_{\pm,\Delta,\zeta}\left[\mathbb I\left(\varphi(\omega_\pm)\neq \pm 1\right)\right]\eqqcolon\epsilon_g.
\end{split}\label{eq:errori}
\end{equation}
\end{result}
\begin{result}
Under the hypothesis that $\mathbb E[\Delta]<+\infty$ and $\mathbb E[\Delta^{-2}]<+\infty$, the Bayes optimal test error for binary classification is
\begin{equation}\comprimi
\epsilon_{\rm g}^{\rm BO}=\rho_+\mathbb E\left[\hat\Phi(\kappa_+)\right]+\rho_-\mathbb E\left[\hat\Phi(\kappa_-)\right],\qquad \kappa_\pm\coloneqq
\frac{1}{\sqrt{\Delta_0^\star \left(1+\frac{\mathbb E[\Delta]\mathbb E[\Delta^{-2}]}{\alpha\mathbb E[\Delta^{-1}]^2}\right)}}\left(1\pm \frac{\Delta_0}{2}\left(1+\frac{1}{\alpha\mathbb E[\Delta^{-1}]}\right)\ln\frac{\rho_+}{\rho_-}\right)
\end{equation}
where the expectation is over the two identically distributed random variables $\Delta_0,\Delta_0^\star\sim\varrho$, whilst  $\hat\Phi(x)$ is the complementary error function.

\end{result}

\paragraph{Quadratic loss with ridge regularisation.}
\label{th:3}
Explicit choices for $\ell$ or $r$ can highly simplify the fixed-point equations above. As an example, let us consider the case of a quadratic loss, $\ell(y,x)=\frac{1}{2}\left(y-x\right)^2$ with ridge regularisation $r(\bw)=\frac{1}{2}\|\bw\|^2$. In this setting, an explicit formula for the proximal $h_\pm$ in Eq.~\eqref{eq:prox_h} and for the proximal $\bgg$ in Eq.~\eqref{eq:prox_g} can be easily found. As a result, we can obtain explicit expressions for the corresponding set of state evolution equations which become particularly suitable for a fast numerical solution and explicit the dependence on higher moments of $\Delta$:
\begin{equation}\label{eq:sp_sq_ridge}
\begin{cases}
m_\pm = \frac{\sum_{k=\pm}\hM_{k}\bmu_{k}^\intercal\bmu_\pm}{\lambda+\hat v},\\
q =\frac{\|\sum_{k=\pm}\hM_{k}\bmu_k\|^2+\hat q}{(\lambda+\hat v)^2},\\
v =\frac{1}{\lambda+\hat v},
\end{cases} \qquad
\begin{cases}
\hat q=
\alpha \mathbb E_\pm[(\pm1-m_\pm-b)^2]\mathbb E_\Delta\left[\frac{\Delta}{(1+v\Delta)^2}\right]+
\alpha q\mathbb E_\Delta\left[\frac{\Delta^2}{(1+v\Delta)^2}\right],\\
\hat v=\alpha\mathbb E_\Delta\left[\frac{\Delta}{1+v\Delta }\right],\\
\hM_\pm=\alpha\rho_\pm(\pm 1-m_\pm-b)\mathbb E_{\Delta}\left[\frac{1}{1+v\Delta}\right].
\end{cases}
\end{equation}

\section{Application to synthetic datasets}\label{sec:examples}

In this section, we compare our theoretical predictions with the results of numerical experiments for a large family of data distributions. The results have been obtained using ridge $\ell_2$-regularisation, and both quadratic and logistic losses, with various data cluster balances $\rho$. We will also assume, without loss of generality, that $\bmu_\pm=\pm\frac{1}{\sqrt d}\bmu$, where $\bmu$ is extracted from a distribution $\mathcal N(\mathbf 0,\bI_d)$. Our artificial data sets will be produced using, for the $\Delta$ distribution $\varrho$, an inverse-Gamma distribution parametrised as
\begin{subequations}\label{eq:invgamma}
\begin{equation}\textstyle
    \varrho(\Delta)\equiv \varrho_{a,c}(\Delta)= \frac{c^a}{\Gamma(a)\Delta^{a+1}}  \e^{-\frac{c}{\Delta}},
\end{equation}
depending on the shape parameter $a>0$ and on the scale parameter $c>0$. As a result, each class will be distributed as
\begin{equation}\textstyle
P(\bx|\bmu)=\frac{(2c)^a \Gamma \left(a+\sfrac{d}{2}\right)}{ \Gamma(a)\pi ^{\sfrac{d}{2}} (2 c+\|\bx-\bmu\|^2)^{a+\sfrac{d}{2}}},\label{eq:invgamma_p}
\end{equation}
\end{subequations}
which produces a cloud centered in $\bmu$. The reason for our choice is that the distribution $\varrho_{a,c}$ allows us to easily explore different regimes by changing two parameters only. Indeed, the distribution has, for $a>1$, covariance matrix $\bSigma=\mathbb E[(\bx-\bmu)\otimes(\bx-\bmu)^\intercal]=\sigma^2\bI_d$, where $\sigma^2=\frac{c}{a-1}$. By fixing $\sigma^2$ and taking the limit $a\to+\infty$, $P(\bx|\bmu)\to \mathcal N(\bx|\bmu,\sigma^2\bI_d)$, i.e., the Gaussian case discussed by \citet{mignacco20a}. For $a\in(0,1]$, instead, $\sigma^2=+\infty$. Note that the choice of inverse-Gamma distribution for $\varrho$ is not that unusual. It has been adopted, for example, to describe non-Gaussian data in quantitative finance \citep{Delpini2011, Langrene_2015} or econometrics models \citep{Nelson1990}. Finally, we will construct our label estimator as in Eq.~\eqref{eq:intro:estimator} with $\varphi(x)=\mathrm{sign}(x)$.

\begin{wrapfigure}{l}{0.45\textwidth}
    \vspace{-0.5cm}\centering
    \includegraphics[width=0.45\textwidth]{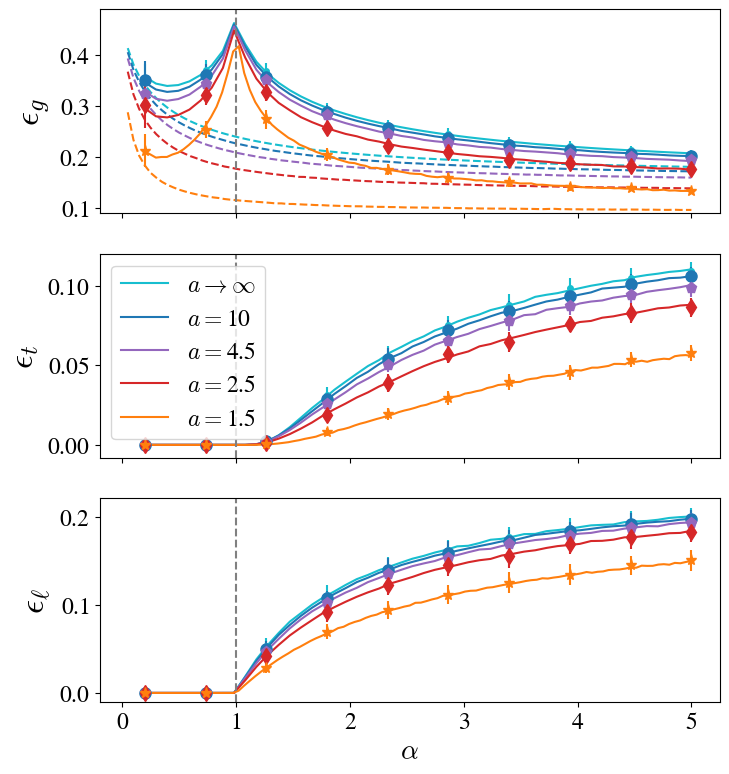}
    \caption{Test error $\epsilon_g$ and (solid line, \textit{top}), training error $\epsilon_t$ (\textit{center}) and training loss $\epsilon_\ell$ (\textit{bottom}) as predicted by Eq.~\eqref{eq:errori} in the balanced $\rho=\nicefrac{1}{2}$ case. The dataset distribution is parametrised as in Eq.~\eqref{eq:invgamma0}. The classification task is solved using a quadratic loss with ridge regularisation with $\lambda=10^{-5}$. In the top figure, the dashed line corresponds to the Bayes optimal bound. Dots correspond to the average outcome of $50$ numerical experiments in dimension $d=10^3$. In our parametrisation, the population covariance is $\bSigma=\bI_d$ for all values of $a$ and moreover, for $a\to+\infty$, the case of Gaussian clouds with the same centroids and covariance is recovered. For further details on the numerical solutions, see Appendix \ref{app:numerics}.}
    \vspace{-0.6cm}
    \label{fig:square_zeroregul}
\end{wrapfigure}

\subsection{Finite-covariance case}\label{sec:examples:gammainv}

Let us start by considering a data set distribution as in Eq.~\eqref{eq:invgamma} with shape parameter $a=c+1> 1$. The data set distribution in this case is given by
\begin{equation}\label{eq:invgamma0}\textstyle
P(\bx|\bmu)=  \frac{2^a (a-1)^a   \Gamma \left(a+\nicefrac{d}{2}\right)}{\Gamma (a)\pi ^{\nicefrac{d}{2}}\left(2 a-2+\|\bx-\bmu\|^2\right)^{a+\nicefrac{d}{2}}}  
\end{equation}

and decays as $\|\bx\|^{-2a-1}$ in the radial direction for $\|\bx-\bmu\|\gg 0$. As a consequence, the distribution has $\lim_d\sfrac{1}{d}\mathbb E[\|\bx-\bmu\|^k]=+\infty$ for $k\geq 2a$. With this choice, \textit{all elements of the family in Eq.~\eqref{eq:invgamma0} have the same covariance $\bSigma=\mathbb E[(\bx-\bmu)\otimes(\bx-\bmu)^\intercal]=\bI_d$ as $a$ is varied}, including the Gaussian limit $P(\bx|\bmu)\to\mathcal N(\bmu,\bI_d)$ obtained for $a\to+\infty$.

In Fig.~\ref{fig:square_zeroregul} we present the results of our numerical experiments using the square loss and small regularisation. An excellent agreement between the theoretical predictions and the results of numerical experiments is found for a range of values of $a$ and sample complexity $\alpha$, both for balanced, i.e., equally sized, and unbalanced clusters of data (the plot for this case can be found in Appendix \ref{app:ridge_regul}). The test error $\epsilon_g$ presents the classical interpolation peak at $\alpha=1$, smoothened by the presence of a non-zero regularisation strength $\lambda$, and the typical double-descent behavior \citep{Opper1990}. As $a$ is increased, the results of \citet{mignacco20a} for Gaussian clouds with $P(\bx|\bmu)=\mathcal N(\bx|\bmu,\bI_d)$ are approached. The $a\to+\infty$ curves correspond, therefore, to the Gaussian mixture model we would have constructed by simply fitting the first and second moments of each class in the finite-$a$ case. The plot shows that, at given population covariance $\bSigma=\bI_d$, the classification of power-law distributed clouds has a different, and in particular smaller, test error than the corresponding task on Gaussian clouds with the same covariance: in this sense, no Gaussian universality principle holds. This (perhaps) counter-intuitive effect stems from the fact that, although for all classes $\bSigma=\bI_d$, the mean absolute deviation is smaller for finite $a$
\begin{equation}\comprimi
\textstyle\lim\limits_{d\to+\infty}\frac{\mathbb E[\|\bx-\bmu\|]}{\sqrt{\mathbb E[\|\bx-\bmu\|^2]}}=\frac{\Gamma(a-\sfrac{1}{2})\sqrt{a-1}}{\Gamma(a)}\xrightarrow{a\to +\infty}1^-.
\end{equation}
The learning process benefits from the fact that points are on average closer to their means $\bmu_\pm$, although the fat tails produce the same variance. Finally, we observed that test errors are systematically smaller for unbalanced clusters than for balanced clusters (see Appendix \ref{app:ridge_regul}).

\begin{wrapfigure}{r}{0.5\textwidth}
    \vspace{-1.2cm}
    \centering
    \includegraphics[width=0.5\textwidth]{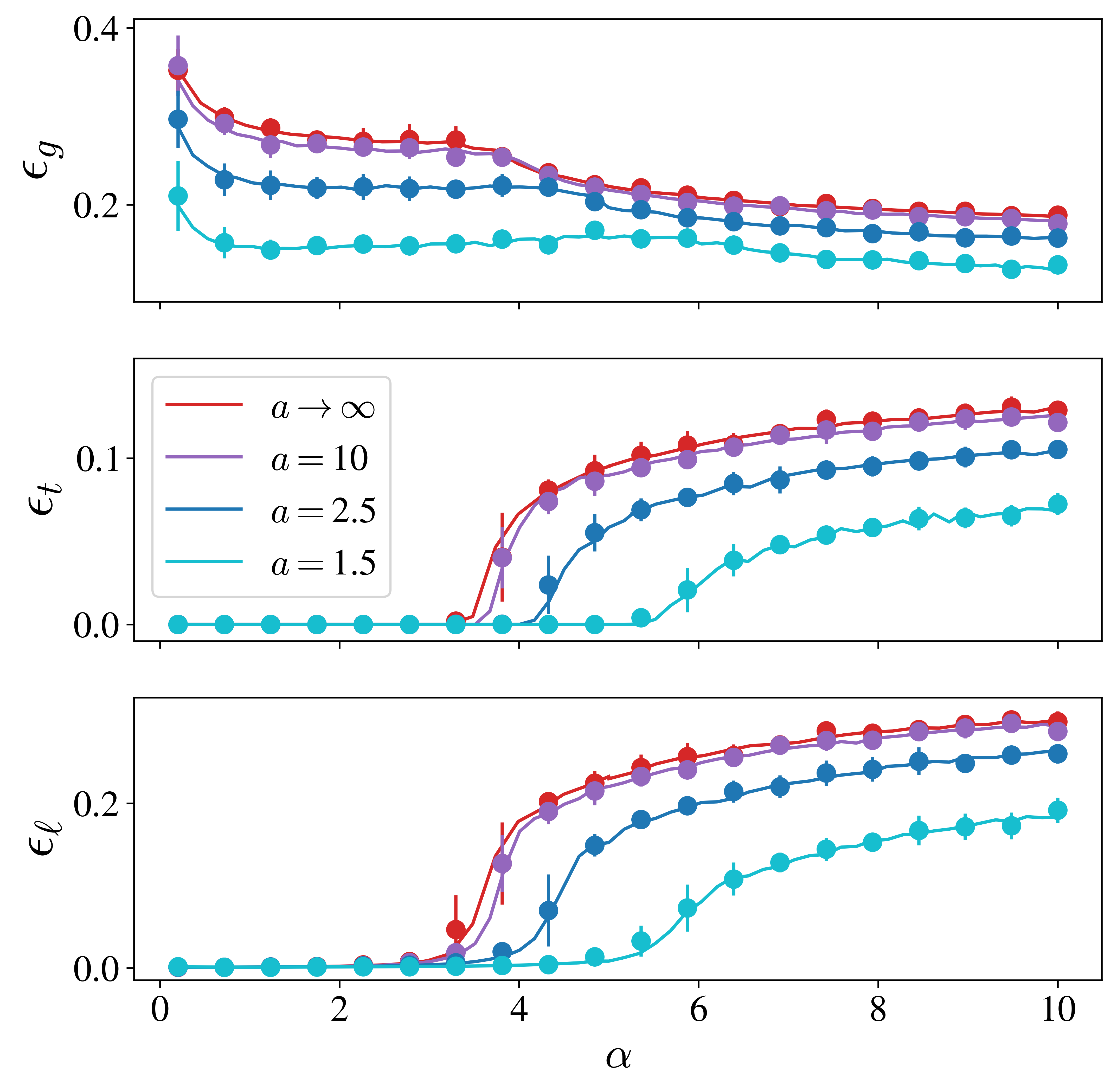}
    \caption{Test error $\epsilon_g$ (\textit{top}), training error $\epsilon_t$ (\textit{center}) and training loss $\epsilon_\ell$ (\textit{bottom}) via logistic loss training on balanced clusters parametrised as in Eq.~\eqref{eq:invgamma0} ($\bSigma=\bI_d$). A ridge regularisation with $\lambda=10^{-4}$ is adopted. Dots correspond to the average over $20$ numerical experiments with $d=10^3$. The Gaussian limit is recovered for $a\to+\infty$. Further details on the numerical solutions can be found in Appendix \ref{app:numerics}.}\vspace{-1cm}
    \label{fig:log}
\end{wrapfigure}

The same numerical experiment has been repeated using the logistic loss for training. Once again, in Fig.~\ref{fig:log}, we focus on $\rho=\nicefrac{1}{2}$ and show that the theoretical predictions of the generalisation and training errors agree with the results of numerical experiments for a range of sample complexity values $\alpha$ and various values of $a$. 
Just as for the square loss, the test error $\epsilon_g$ of the Gaussian model ($a\to+\infty$) is larger than the one observed for power-law distributed clouds at finite values of $a$. In the $\lambda\to 0$ limit, the typical interpolation cusp in the generalisation error is observed: the cusp occurs at different values of $\alpha$ as $a$ is changed; interpolation occurs at smaller values of $\alpha$ for larger $a$ (i.e., for ``more Gaussian'' distributions). A comparison with the test error $\epsilon_t$ confirms that this cusp occurs at the value of $\alpha$ where the training error becomes non-zero and the two training data clouds become non-separable \citep{SurCandes2019, Deng2021}. This sharp transition in $\alpha$ also corresponds to the existence transition of maximum-likelihood estimator in high-dimensional logistic regression, analysed for Gaussian data in Refs.~\citep{SurCandes2019,mignacco20a} and, in our setting, in Section~\ref{sec:sepa} below. For larger regularisation strength, the cusp smoothens, and the training error becomes non-zero at smaller values of $\alpha$. 

\subsection{Infinite-covariance case}
\begin{figure}[b!]
    \centering
    \includegraphics[width=0.45\textwidth]{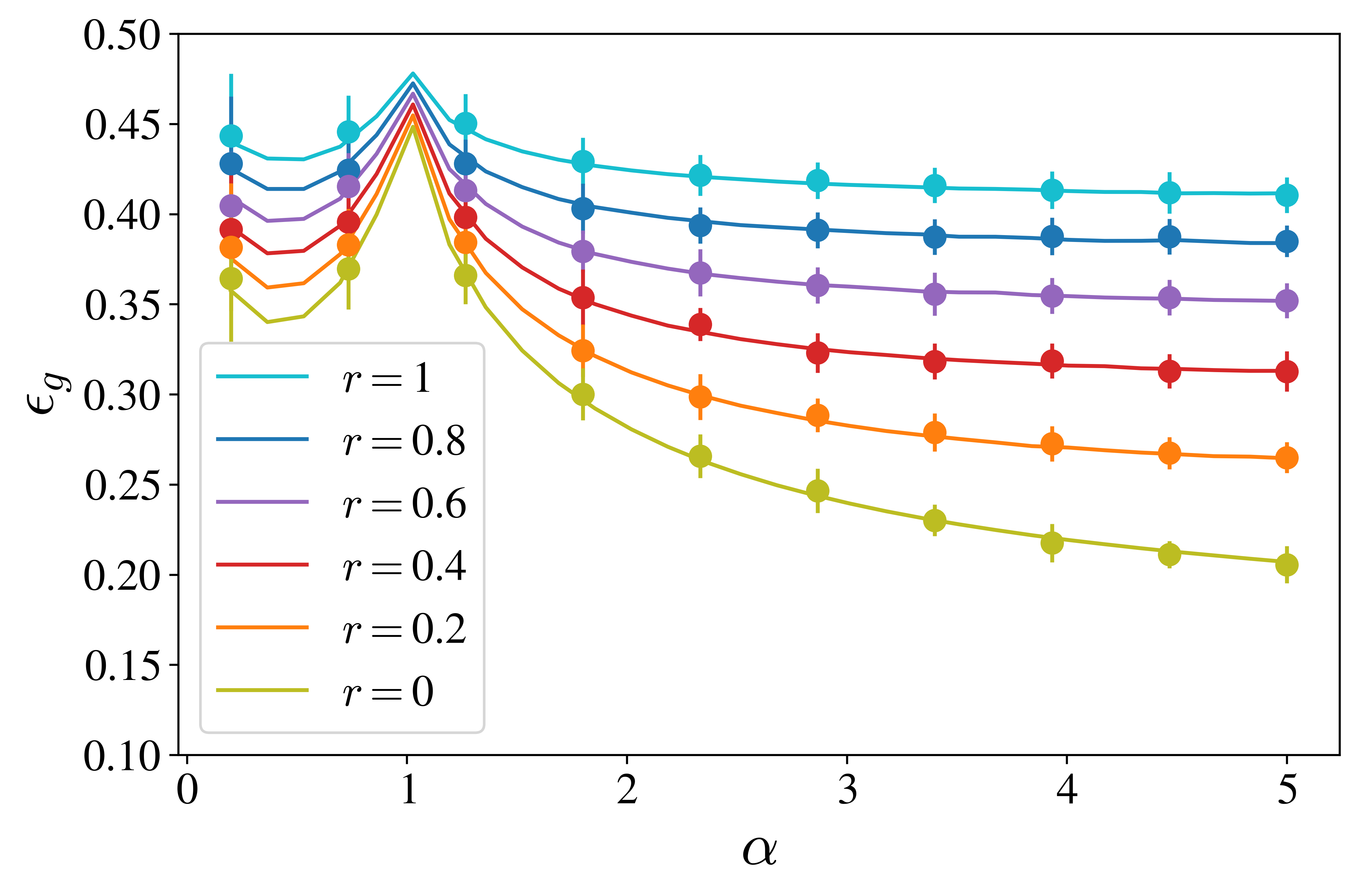}
    \hfill
    \includegraphics[width=0.45\textwidth]{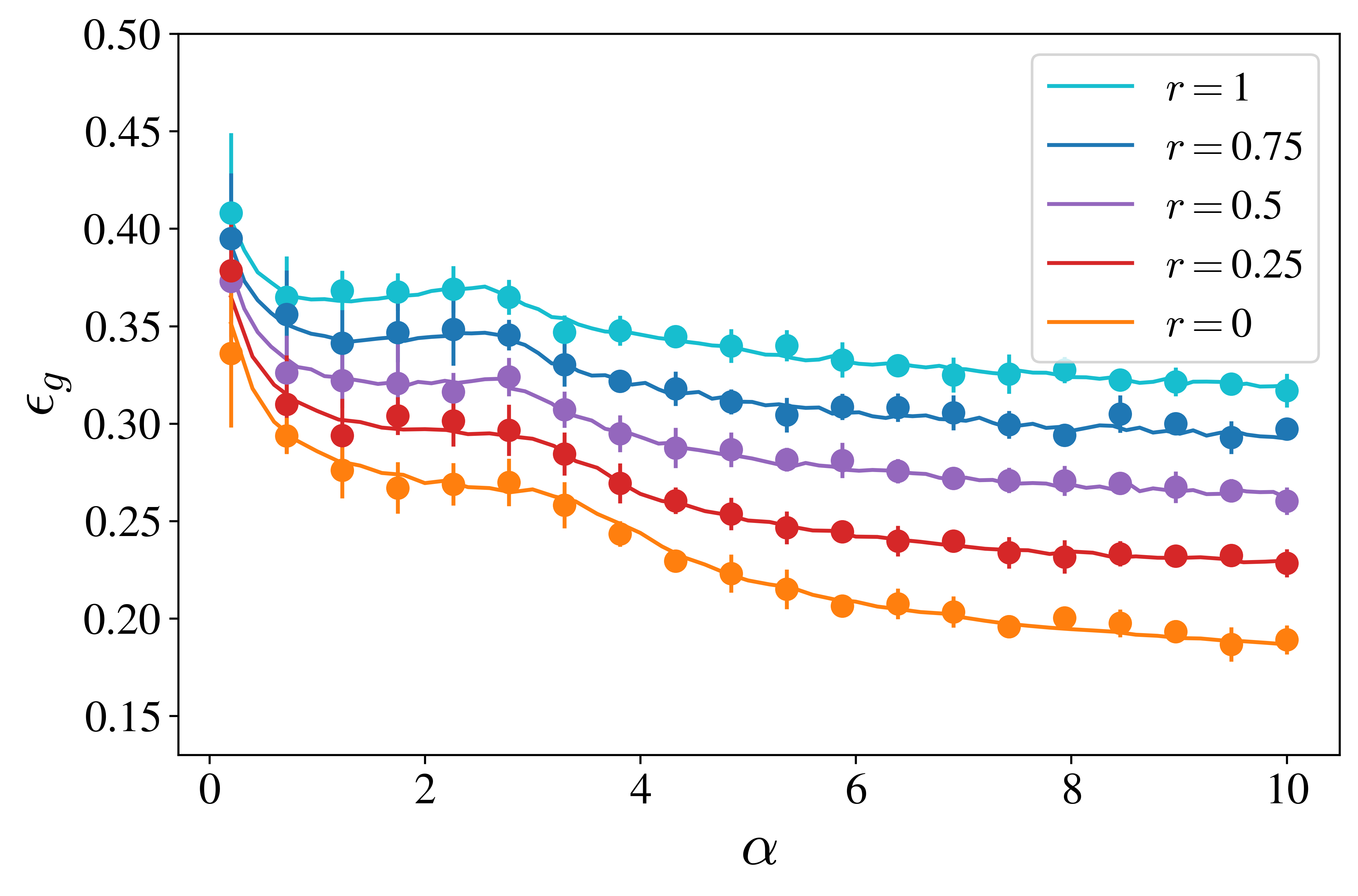}
    \caption{Test error $\epsilon_g$ in the classification of two balanced clouds, via quadratic loss (\textit{left}) and logistic loss (\textit{right}). In both cases, ridge regularisation is adopted ($\lambda=10^{-4}$ for the square loss case, $\lambda=10^{-3}$ for the logistic loss case). Each cloud is a superposition of a power-law distribution with infinite variance and a Gaussian with covariance $\bSigma=\bI_d$. The parameter $r$ allows us to contaminate the purely Gaussian case ($r=0$) with an infinite-variance contribution ($0<r\leq 1$) as in Eq.~\eqref{eq:invgamma} with $c=1$ and $a=\nicefrac{1}{2}$ (\textit{left}) or $a=\nicefrac{3}{4}$ (\textit{right}). Dots correspond to the average test error of $20$ numerical experiments in dimension $d=10^3$. Note that, at a given sample complexity, Gaussian clouds are associated with the lowest test error for both losses.}\label{fig:eg_iterp}
\end{figure}
The family of distributions specified by Eq.~\eqref{eq:invgamma} allows us to also consider the case of infinite covariance, i.e., $\sigma^2=+\infty$. To test our formulas in this setting, we considered the case in which each cloud is obtained by a ``contaminating'' a Gaussian with an infinite-covariance distribution as in Eq.~\eqref{eq:invgamma_p} with $c=1$ and $a<1$ \cite{Huber1964}. In other words, we adopted as dataset distribution the density $\varrho(\Delta)=r\varrho_{a,1}(\Delta)+(1-r)\delta(\Delta-1)$, with $r\in[0,1]$ interpolation parameter, for $\Delta$ in Eq.~\eqref{mixture}. Each class has therefore infinite covariance for $0<r\leq 1$ and the Gaussian limit is recovered for $r=0$. Fig.~\ref{fig:eg_iterp} collects our results for two balanced clouds, analysed using square loss and logistic loss. Good agreement between the theoretical predictions and the results of numerical experiments is found for a range of values of sample complexity and of the ratio $r$. In this case, the finite-variance case $r=0$ corresponds, not surprisingly, to the lowest test error with both square loss and logistic loss.

\subsection{The role of regularisation in the classification of non-Gaussian clouds}
\begin{figure}[!htb]
 \centering
 \includegraphics[width=0.32\textwidth]{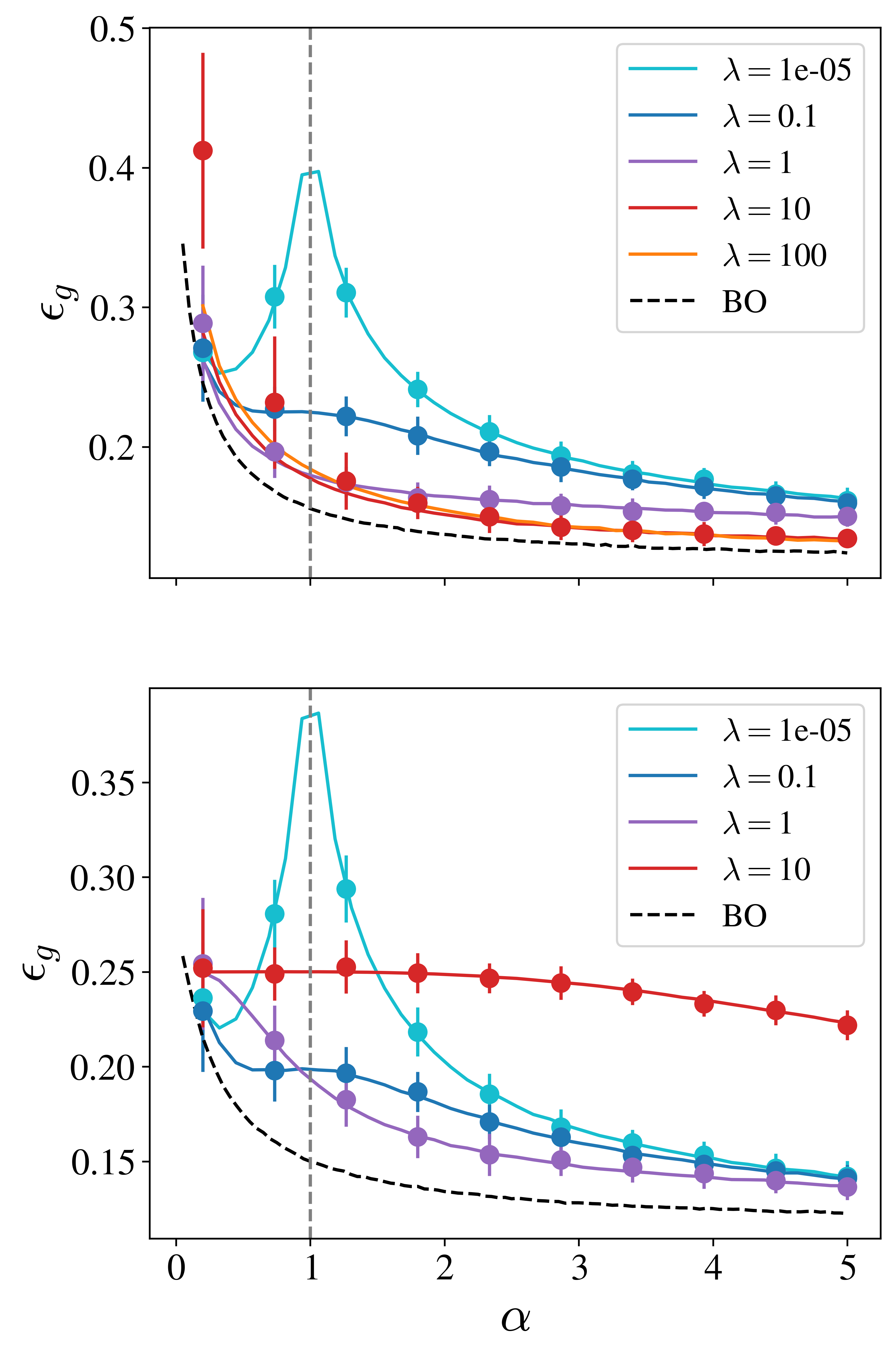}
 \hfill
 \includegraphics[width=0.32\textwidth]{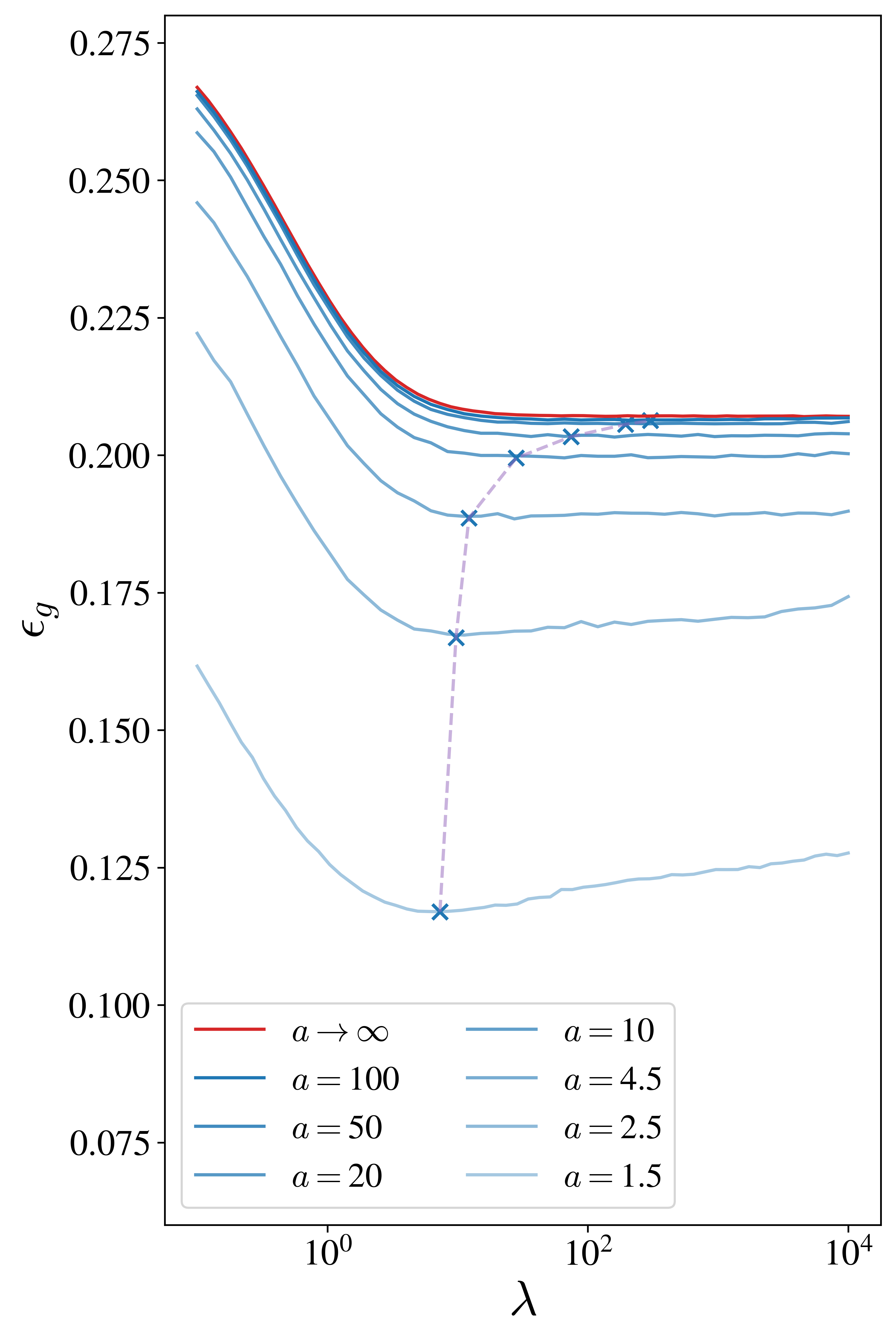}
 \hfill
 \includegraphics[width=0.32\textwidth]{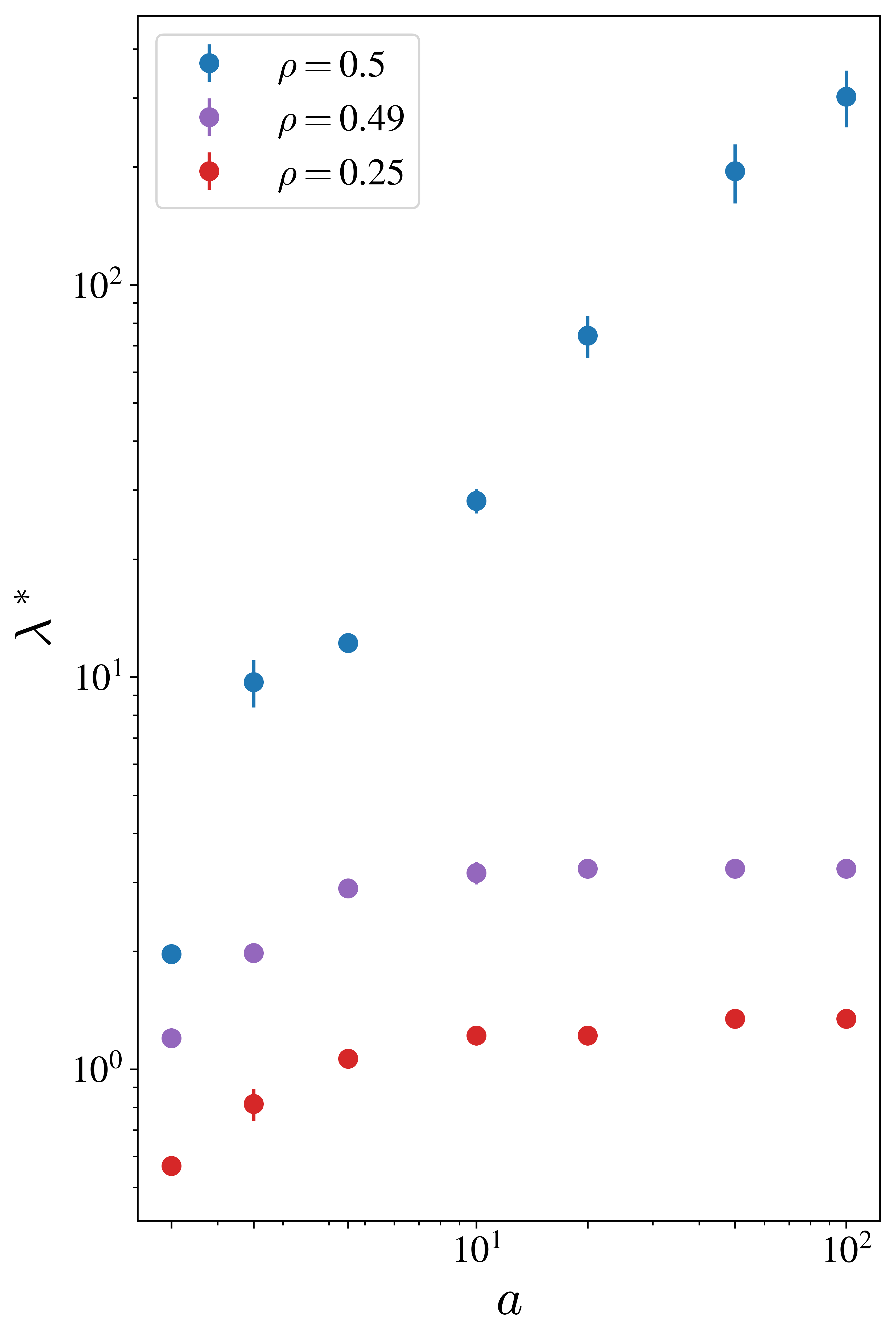}
    \caption{(\textit{Left}) Test error for ridge regularised quadratic loss for various regularisation strengths. The data points of each cloud in the training set are distributed as in Eq.~\eqref{eq:invgamma0}, with shape parameter $a=2$, for balanced clusters (\textit{top}) and unbalanced clusters ($\rho=\nicefrac{1}{4}$, \textit{bottom}). Points are the results of $50$ numerical experiments, and the dashed lines are Bayes-optimal bounds. (\textit{Center}) Test error for different regularisation strengths $\lambda$ for two \textit{balanced} clusters with quadratic loss at sample complexity $\alpha=2$ using the data distribution \eqref{eq:invgamma0}. The optimal regularisation strength value $\lambda^\star$ obtained from averaging 5 runs for each $a$ is marked with a cross. (\textit{Right}) Optimal regularisation strength $\lambda^\star$ at $\alpha=2$ for different values of $a\in[1.5,10^2]$ for both balanced and unbalanced clusters, obtained from averaging 5 runs. Note that, for $\rho=\nicefrac{1}{2}$, $\lambda^\star\to+\infty$ as $a\to+\infty$. \label{fig:images}}
\end{figure}

One of the main results in the work of \citet{mignacco20a} is related to the effect of regularisation on the performances in a classification task on a Gaussian mixture. They observed that, for all values of the sample complexity $\alpha$, the optimal ridge classification performance on a pair of balanced clouds is obtained for an infinite regularisation strength $\lambda$, using both hinge loss and square loss. We tested the robustness of this result beyond the purely Gaussian setting. In Fig.~\ref{fig:images} (left), we plot the test error obtained using square loss and different regularisation strengths $\lambda$ on the dataset distribution as in Eq.~\eqref{eq:invgamma0} with $a=2$. We observe that, both in the balanced case and in the unbalanced case, the optimal regularisation strength $\lambda^\star$ is \textit{finite} and, moreover, $\alpha$-dependent. On the other hand, if we fix $\alpha$ and let $a$ grow towards $+\infty$, the optimal $\lambda^\star$ grows as well to recover the result of \citet{mignacco20a} of diverging optimal regularisation for \textit{balanced} Gaussian clouds, as shown in Fig.~\ref{fig:images} (center) for $\alpha=2$. 
This result is represented also in Fig.~\ref{fig:images} (right), where it is shown that, in the \textit{unbalanced} case, the optimal regularisation strength $\lambda^\star$ instead saturates for $a\to+\infty$ to a finite value, as observed for Gaussian clouds \citep{mignacco20a}.

\begin{figure}[t]
    \centering
    \includegraphics[height=0.35\textwidth]{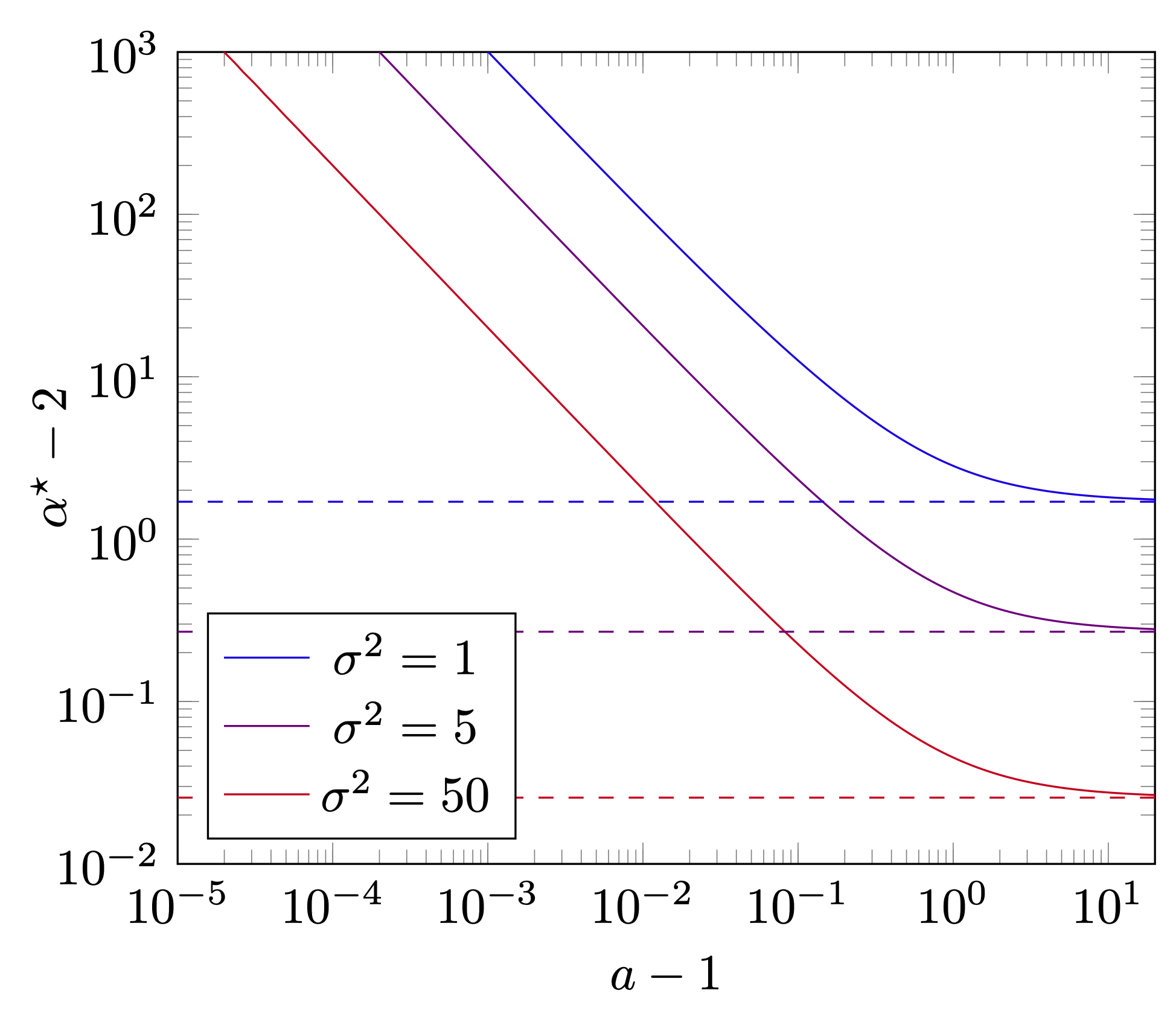}\qquad
    \includegraphics[height=0.35\textwidth]{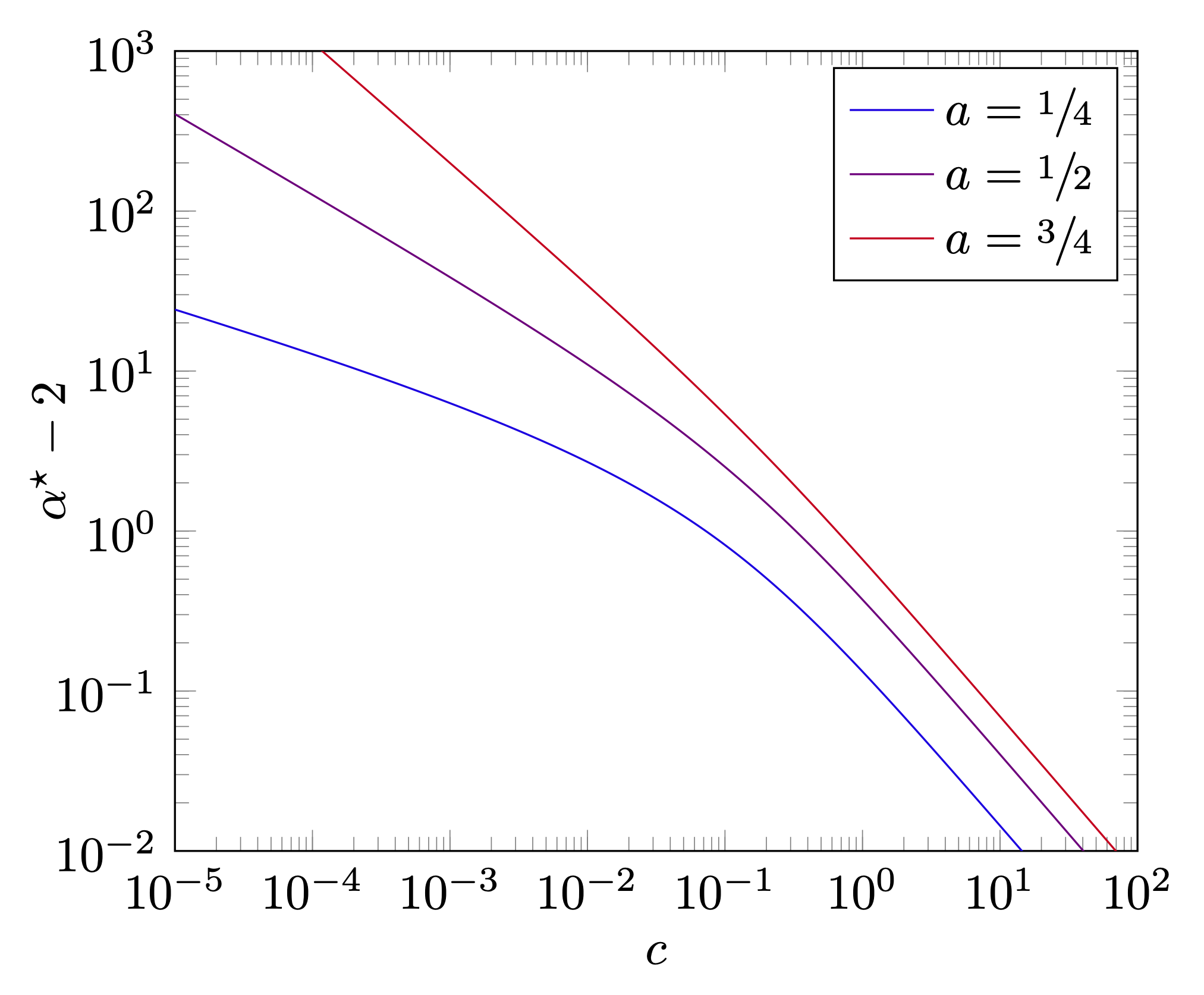}
    \caption{Separability threshold $\alpha^\star$ obtained by solving the equations in Eq.~\eqref{spgeneral} with logistic loss, ridge regularisation strength $\lambda=10^{-5}$ and $\rho=\nicefrac{1}{2}$. The data points of each cloud are distributed around their mean $\bmu$ with a two-parameter distribution as in Eq.~\eqref{eq:invgamma_p}. (\textit{Left}). Finite covariance $\bSigma=\sigma^2\bI_d$ case, {$\sigma^2=\frac{c}{a-1}$}, as a function of $a$. Dashed lines are the threshold values of the Gaussian case derived by \citet{mignacco20a}. At large $a$ and large variance $\sigma^2$, the Cover's transition $\alpha=2$ for balanced clusters is recovered. (\textit{Right}) Infinite-variance data clusters case. The cluster distribution is obtained by fixing $0<a<1$ in Eq.~\eqref{eq:invgamma_p}. 
    }
    \label{fig:separability_thresholds}
\end{figure}

\section{The separability threshold}
\label{sec:sepa}
By studying the training error $\epsilon_t$ with logistic loss at zero regularisation we can obtain information on the boundary between the region in which the training data are perfectly separable and the regime in which they are not. In other words, we can extract the value of the so-called separability transition complexity $\alpha^\star$ \citep{SurCandes2019,mignacco20a,loureiro2021}. Once again, a complete characterisation of this transition point is available in the Gaussian mixture case, where $\alpha^\star$ can be explicitly given as a function of $\sigma^2$ \citep{mignacco20a}. 

It is not trivial to extend such a result to the general case of non-Gaussian mixtures. It is however possible to estimate $\alpha^\star$ within the large family of ``superstatistical'' models we are considering here under the assumption that Result \ref{th:claim} holds. In this case, in Appendix \ref{app:logistic} we prove the following.
\begin{result}
In the considered double-stochastic model, data are linearly separable for $\alpha<\alpha^\star$, where
\begin{equation}\comprimi
\alpha^\star\coloneqq\max_{\theta\in(0,1],\gamma}\frac{1-\theta^2}{\mathcal S(\theta,\gamma)},\qquad \mathcal S(\theta,\gamma)\coloneqq \int_0^\infty f^2\mathbb E_\Delta\left[\rho_+\mathcal N\left(f+\frac{\theta+\gamma}{\sqrt{\Delta}}\big|0,1\right)+\rho_-\mathcal N\left(f+\frac{\theta-\gamma}{\sqrt{\Delta}}\big|0,1\right)\right]\dd f.\label{eq:separa}
\end{equation}
\end{result}
Fig.~\ref{fig:separability_thresholds} shows the values of $\alpha^\star$ for different choices of the shape parameters $a$ and $c$ in the case of two balanced clouds of datapoints distributed as in Eq.~\eqref{eq:invgamma_p} as predicted by Eq.~\eqref{eq:separa}. In Fig.~\ref{fig:separability_thresholds} (left), we fixed  $\sigma^2=\frac{c}{a-1}<+\infty$, and plotted the separability threshold $\alpha^\star$ for a range of values of the shape parameter $a>1$. As $a$ grows, the known threshold value for the Gaussian mixture case is recovered \citep{mignacco20a}. The double limit $a\rightarrow\infty$ and $\sigma^2\rightarrow\infty$, on the other hand, provides the \citet{Cover1965} transition value $\alpha^\star = 2$, as expected. In Fig.~\ref{fig:separability_thresholds} (right), instead, we analyse the case of infinite-variance clusters, by fixing $a\in(0,1)$ and by varying the scale parameter $c$ in Eq.~\eqref{eq:invgamma_p}. In this setting, $c$ plays the role of distribution width. The Cover transition is therefore correctly recovered as $c$ diverges for all values of $a$.

\section{Random labels in the teacher-student scenario and Gaussian universality} 

Our analysis can be extended to the case in which a point $\bx$ is labeled not according to its class but by a ``teacher'', represented by a probabilistic law $P_0(y|\tau)$ and a vector $\btheta_0\in\R^d$ so that each dataset element is independently generated with the joint distribution
\begin{equation}\textstyle
P(\bx,y)=P_0\left(y\Big|\frac{\btheta_0^\intercal\bx}{\sqrt d}\right)\sum\limits_{k=\pm}\rho_k\mathbb E[\mathcal{N}\left(\bx\left|\bmu_k,\Delta\bI_d\right.\right)].    
\end{equation}
Let us assume now that $\btheta_0$ is such that $\lim_d\sfrac{1}{\sqrt d}\btheta^\intercal_0\bmu_\pm= 0$, i.e., the teacher is ``uncorrelated'' with the structure of the mixture. In Appendix \ref{App:RL} we show that, if $\ell(y,\eta)=\ell(-y,-\eta)$ and $P_0(y|\tau)=P_0(-y|-\tau)$, the task is equivalent to a classification problem on a \textit{single} cloud centered in the origin, i.e., to the case $\bmu_\pm=\mathbf 0$, so that $P(\bx,y)=P_0\big(y\big|\sfrac{1}{\sqrt d}\btheta_0^\intercal\bx\big)\mathbb E[\mathcal{N}\big(\bx\big|\mathbf 0,\Delta\bI_d\big)]$. This mean-invariance has been recently proven in Ref.~\cite{gerace2023} in the pure Gaussian setting and it holds therefore to a non-Gaussian mixture.

A special, relevant case of ``uncorrelated teacher'' is the one of \textit{random labels}, where $P_0(y|\tau)=\sfrac{1}{2}(\delta_{y,+1}+\delta_{y,-1})$. This apparently vacuous model is relevant in the study of the storage capacity problem \citep{Gardner_1989_unfinished,KrauthMezard89_cap,Brunel1992} and of the separability threshold \citep{Cover1965}, and it has also been an effective tool to better understand deep learning \citep{Maennel2020,Zhang2021}. In this case, the following universal result holds in our setting.
\begin{result}
In the random label setting studied via ridge regression case with zero regularisation, the asymptotic training loss is independent of the distribution of $\Delta$, and given by the universal formula
\begin{equation}\label{eq:eluni}\comprimi
\frac{1}{2n}\sum_{\nu=1}^\infty\left(y^\nu-\frac{\bx^\intercal\bw^\star}{\sqrt d}\right)^2\xrightarrow[\sfrac{n}{d}=\alpha]{n\to+\infty}\frac{\alpha-1}{2\alpha}\theta(\alpha-1)\eqqcolon\epsilon_\ell.
\end{equation}
\end{result}
The proof of the previous result is given in Appendix \ref{App:RL}. Note that this result does not require $\sigma^2\coloneqq\mathbb E[\Delta]$ to be finite, and it includes, as a special case, the Gaussian mixture model with random labels considered in Ref.~\cite{gerace2023,pesce2023}. Therein, Eq.~\eqref{eq:eluni} has been obtained under the Gaussian design hypothesis. It turns out therefore that Eq.~\eqref{eq:eluni} holds for a much larger ensemble, which includes fat-tailed distributions with possibly infinite covariance. On the other hand, universality \textit{does not} hold, e.g., for the training error. 

\begin{wrapfigure}{r}{0.45\textwidth}
    \centering
    \vspace{-1cm}\includegraphics[width=0.45\textwidth]{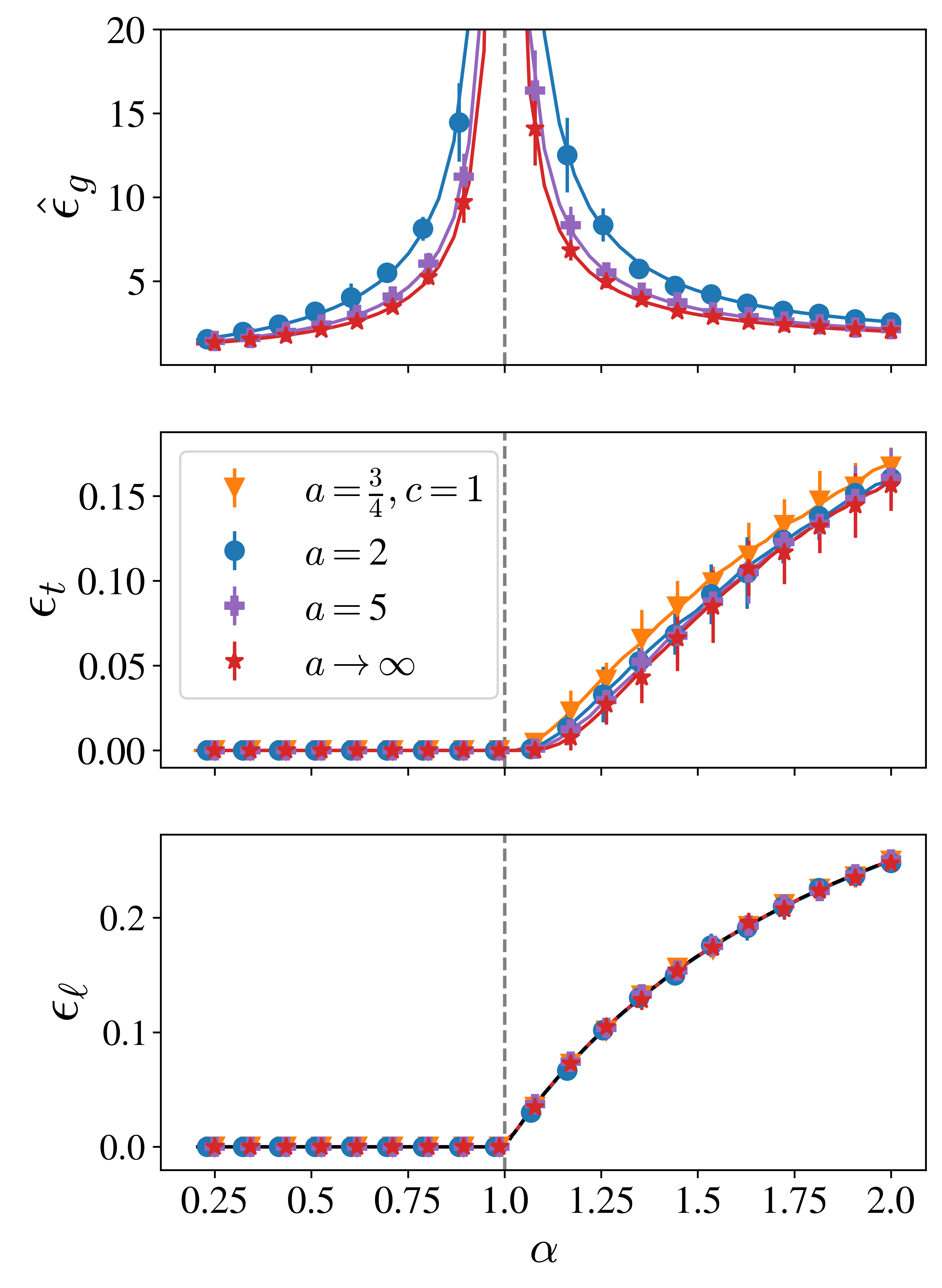}
    \caption{Mean square error $\hat\epsilon_g$ (\textit{top}), training error $\epsilon_t$ (\textit{center}) and training loss $\epsilon_\ell$ (\textit{bottom}) obtained using square loss on a dataset distributed as in Eq~\eqref{eq:datarl} with random labels (dots) compared with the theoretical predictions for the single cluster case (continuous lines). The distributions are parametrised using Eq.~\eqref{eq:invgamma0} for $a>1$ and Eq.~\eqref{eq:invgamma_p} for $0<a\leq 1$. A ridge regularisation with $\lambda=10^{-4}$ is adopted. Note that $\hat\epsilon_g$ is absent in the top plot for $a=\sfrac{1}{2}$ and $c=1$ being a divergent quantity in this case. }\vspace{-0.5cm}
    \label{fig:unirl}
\end{wrapfigure}

In Fig.~\ref{fig:unirl} we exemplify these results by using ridge-regularised square loss. The numerical experiments correspond to a classification task on a dataset of \textit{two} clouds centered in $\bmu_\pm$, $\bmu_+=-\bmu_-\sim\mathcal N(\mathbf 0,\sfrac{1}{d}\bI_d)$, obtained using the distribution
\begin{equation}\label{eq:datarl}\textstyle
P(\bx,y)=\frac{\delta_{y,+1}+\delta_{y,-1}}{4}\sum\limits_{k=\pm}P(\bx|\bmu_k),  
\end{equation}
where $P(\bx|\bmu_\pm)$ is the distribution in Eq.~\eqref{eq:invgamma_p} depending on the shape parameters $a$ and $c$. These results are compared with the theoretical prediction obtained for a \textit{single} cloud, with the same parameters $a$ and $c$, but centered in the origin, so that $P(\bx,y)=\frac{\delta_{y,+1}+\delta_{y,-1}}{2}P(\bx|\mathbf 0)$: the agreement confirms the ``mean-invariance'' property stated above. As anticipated, as we are considering random labels, the training loss $\epsilon_\ell$ follows Eq.~\eqref{eq:eluni} in all cases, whereas the test error is not universal. Note that, in place of the test error (which is trivially $\epsilon_g=\sfrac{1}{2}$) we plot the mean square error $\hat\epsilon_g\coloneqq \mathbb E_{(y,\bx)}[(y-\sfrac{1}{\sqrt d}\bx^\intercal\bw^\star)^2]$. Also, observe that this quantity is divergent in the case of infinite covariance, hence is absent from the plot for $a\in(0,1]$.

\section{Conclusions and perspectives}\label{sec:conclusion}
The introduced framework has allowed us to obtain an exact asymptotic analysis of the learning of a mixture of two clouds of points with possibly power-law tails. In our dataset, each sample point $\bx$ is labeled by $y\in\{-1,1\}$ and obtained as $\bx=y\bmu+\sqrt{\Delta}\bz$, where $\bmu\in\R^d$, $\bz\sim\mathcal N(\mathbf 0,\bI_d)$ and $\Delta\sim\varrho$. We have obtained the exact expression for the test and training error achieved by a GLM in the limit of large size $n$ of the dataset and large dimensionality $d$ under the assumption that $\alpha=\sfrac{n}{d}$ is kept fixed. We have shown that the performance crucially depends on higher-order moments of the distributions, so that a Gaussian assumption is not feasible. Moreover, we have shown the effects of non-Gaussianity on the optimal regularisation strength and derived an exact formula for the critical $\alpha^\star$ at which the two clouds become linearly inseparable. Finally, we have considered the so-called random labels setting (in which sample labels are assigned in a completely random way) and shown that, in this case, the training square loss takes a universal asymptotic form independent of the distribution of $\Delta$: this result further supports the crucial role of labels' randomness in some recent universality results \cite{gerace2023}. Our contribution exemplifies the fact that a deviation from Gaussianity in high dimensionality cannot, in general, be neglected by providing an exactly solvable toy model exhibiting power-law tails and possibly unbounded covariance.

The application of the introduced theoretical framework to real datasets is an interesting direction to explore in the future: the main difficulty, in this case, is the choice of the best distribution $\varrho$ given the observed dataset, a problem that however has a long tradition in the context of Bayesian estimation \cite{Alspach,gelman2013bayesian}\footnote{In a more simplistic approach, it can be observed that, in the case in which the square loss is adopted, equations \eqref{eq:sp_sq_ridge} depend on $\mathbb E[(1+v\Delta)^{-1}]$ and $\mathbb E[(1+v\Delta)^{-2}]$ only. Such quantities can be numerically estimated from the dataset, for example, by empirically evaluating the identity $\mathbb E[(1+v\Delta)^{-1}]=\sqrt v\partial_v\int_0^v(v-u)^{-1/2}\mathbb E[\exp\big(-\sfrac{ux^2}{2}\big)]\dd u$ when it is assumed that $x\sim\mathbb E[\mathcal N(0,\Delta)]$.}.

\subsubsection*{Acknowledgements} The authors would like to thank Nikolas N\"usken for stimulating discussions. Also, we would like to thank Bruno Loureiro for his feedbacks and for his careful reading of the manuscript.

\bibliographystyle{plainnat}
\bibliography{arxov_superst.bib}

\newpage

\newpage
\appendix
\onecolumn
\section{Derivation of the fixed-point equations}
\label{app:replica}
In this Appendix, we will derive the fixed-point equations for the order parameters presented in the main text, following and generalising the analysis in Ref.~\cite{loureiro2021}. The problem will be given in a slightly more general setting then the one considered in the main text, namely considering a multiclass classification task on $K$ classes. The dataset $\mathcal D\coloneqq \{(\bx^\nu,y^\nu)\}_{\nu\in[n]}$ we are going to consider consists of $n$ independent datapoints $\bx^\nu\in\R^d$ each associated with a label $y^\nu\in\mathcal Y\subseteq\R$, where $\mathcal Y=\{y_k\}_{k\in[K]}$ is a finite set of $K$ elements (for example, for $K=2$ it is standard to choose $\mathcal Y=\{-1,1\}$). The elements of the dataset are independently generated by using a law $P(\bx,y)$ which we assume can be put in the form $P(\bx,y)\equiv \int_0^\infty P(\bx,y|\Delta)\varrho(\Delta)\dd\Delta$ for some given distribution on the positive real axis $\varrho$, and such that
\begin{equation}
 P(\bx,k|\Delta)
 =\rho_k\mathcal{N}\left(\bx\left|\bmu_k,\Delta\bI_d\right.\right),\qquad \rho_k\in(0,1)\ \forall k\in\mathcal Y,\quad \sum_k\rho_k=1.
\end{equation}
The vectors $\bmu_k\in\R^d$ play the role of centroids of each cluster $y=k\in\mathcal Y$. We will perform our classification task searching for a set of parameters $(\bw^\star,b^\star)$, called respectively \textit{weights} and \textit{bias}, that will allow us to construct an estimator via a certain classifier
\begin{equation}
\bx\mapsto \varphi\left(\frac{\bw^\star{}^\intercal\bx}{\sqrt d}+b^\star\right)\in\mathcal Y
\end{equation}
to estimate the label of a new, unobserved datapoint $\bx$. Here $\varphi\colon\R\to\mathcal Y$. To fix the ideas, in all our numerical experiments we used $K=2$ and $\varphi(x)=\mathrm{sign}(x)$. The choice of the parameters is performed by minimising an empirical risk function constructed via a loss function $\ell\colon \mathcal Y\times\R\to\R$ and a proper regularisation $r\colon\R^d\to\R$, in the form
\begin{equation}
\mathcal R(\bw,b)\equiv \sum_{\nu=1}^n\ell\left(y^\nu,\frac{\bw^\intercal\bx^\nu}{\sqrt d}+b\right)+\lambda r(\bw),
\end{equation}
i.e., they are given by
\begin{equation}
(\bw^{\star},b^{\star}) \equiv \arg\min_{\substack{\bw\in \R^{d}\\b \in \R}}\mathcal R(\bw,b).  
\end{equation}
We will assume that the loss function $\ell$ is convex with respect to its second argument. The parameter $\lambda\geq0$ tunes the strength of the regularisation $r$, which also is assumed to be convex. The starting point of our approach is the reformulation of the task as an optimisation problem by introducing a Gibbs measure over the parameters $(\bw,b)$ depending on a non-negative parameter $\beta$,
\begin{equation}
\mu_{\beta}(\bw,b)\propto \e^{-\beta\mathcal R(\bw,b)}=
\underbrace{\e^{-\beta r(\bw)}}_{P_w(\bw)}\prod\limits_{\nu=1}^{n}\underbrace{\exp\left[-\beta\ell\left(y^{\nu},\frac{\bw^\intercal\bx^\nu}{\sqrt d}+b\right)\right]}_{P_{y}(y|\bw,b)},
\end{equation}
so that, in the $\beta\to+\infty$ limit, $\mu_{\beta}(\bw,b)$ concentrates on the values $(\bw^\star,b^\star)$ that minimize the empirical risk $\mathcal R(\bw,b)$ and are therefore the goal of the learning process. The functions $P_{y}$ and $P_{w}$ can be interpreted as (unnormalised) likelihood and prior distribution respectively. Our analysis will go through the computation of the average free energy density associated with this Gibbs measure in a specific proportional limit, i.e.,
\begin{equation}\label{eq:replicaf}
f_\beta\coloneqq-\lim_{\substack{n,d\to+\infty\\\nicefrac{n}{d}=\alpha}}\mathbb E_{\mathcal D}\left[\frac{\ln \mathcal Z_\beta}{d\beta}\right]=\lim_{\substack{n,d\to+\infty\\\nicefrac{n}{d}=\alpha}}\lim_{s\to 0}\frac{\ln\mathbb E_{\mathcal D}[\mathcal Z_\beta^s]}{sd\beta},\qquad \mathcal Z_\beta\coloneqq \int \e^{-\beta\mathcal R(\bw,b)}\dd\bw.
\end{equation}
where $\mathbb E_{\mathcal D}[\bullet]$ is the average over the training dataset. To perform the computation of such quantity, we use the so-called replica method.
\subsection{Replica approach}
We proceed in our calculation by assuming no prior on $b$, which will play a role of an extra (low-dimensional) parameter whose optimal value will be derived extremising with respect to it the final result for the free energy. We need to evaluate
\begin{equation}
\mathbb E_{\mathcal D}[\mathcal Z_\beta^s]=\prod_{a=1}^s\int \dd \bw^a P_w(\bw^a)\left(\sum_{k} \rho_k\mathbb E_{\bx|k}\left[\prod_{a=1}^s P_k\left(\frac{\bw^a\bx}{\sqrt d}+b\right)\right]\right)^n.
\end{equation}
Here and in the following, for the sake of brevity, $P_k(\eta)\equiv P_y(y_k|\eta)$ and similarly $\ell_k(\eta)\equiv \ell(y_k|\eta)$. Let us take the inner average introducing a new set of variables $\eta^a$,
\begin{multline}
\mathbb E_{\bx|k}\left[\prod_{a=1}^s P_k\left(\frac{\bw^{a\intercal}\bx}{\sqrt d}+b\right)\right]=\prod_{a=1}^s\int\dd\eta^a P_k(\eta^a)\mathbb E_{\bx|k}\left[\prod_{a=1}^s\delta\left(\eta^a-\frac{\bw^{a\intercal}\bx}{\sqrt d}+b\right)\right]\\
=\mathbb E_{\Delta|k}\left[\prod_{a=1}^s\int\dd\eta^a P_k(\eta^a)\mathcal N\left(\bEta\Big|\frac{\bw^{a\intercal}\bmu_k}{\sqrt d}-b;\frac{\Delta\bw^a{\bw^b}^\intercal}{d}\right)\right],
\end{multline}
where we used the superstatistical form of the distribution of a datapoint $\bx$. Using the shorthand $\mathbb E_{k,\Delta}[
\Phi_k(\Delta)]\equiv \sum_{k}\rho_k\int_0^\infty\varrho(\Delta) \Phi_k(\Delta)\dd\Delta$,
we can write then
\begin{multline}
\mathbb E_{\mathcal D}[\mathcal Z_\beta^s]
=\prod_{a=1}^s\int \dd \bw^a P_w(\bw^a)\left(\mathbb E_{k,\Delta}\left[\prod_{a=1}^s\int\dd\eta^a P_k(\eta^a)\mathcal N\left(\bEta;\frac{\bw^{a\intercal}\bmu_k}{d}+b;\frac{\Delta \bw^a{\bw^b}^\intercal}{d}\right)\right]\right)^n
\\
=\left(\prod_{a\leq b}\iint\mathcal D{\bQ}^{ab}\mathcal D\hat{\bQ}^{ab}\right)
\left(\prod_{a}\iint\mathcal D{\bM}^{a}\mathcal D{\hat \bM}^{a}\right)\e^{-d\beta\Phi^{(s)}}.
\end{multline}
In the equation above we introduced the \textit{order parameters}
\begin{align}
Q^{ab}_\Delta&=\Delta\frac{\bw^{a\top}\bw^{b}}{d}\in \R,\quad a,b=1,\dots, s,\\
m^{a}_k&=\frac{\bw^{a\intercal}\bmu_k}{\sqrt d}\in \R,\quad a=1,\dots,s,\quad k\in\mathcal Y,
\end{align}
whilst $\iint\mathcal D{\bQ}^{ab}\mathcal D{\bhQ}^{ab}$ express the integration over ${Q}^{ab}_\Delta$ and ${\hQ}_\Delta^{ab}$, to be intended as functions of $\Delta$. Similarly, $\mathcal D{m}^{a}\mathcal D{\hat m}^{a}\propto\prod_{k}\dd m_k^a\dd\hat{m}_k^a$. We have also introduced the replicated free-energy
\begin{multline}
\beta \Phi^{(s)}(Q,M,\hQ,\hM,b)=\sum_{k}\sum_{a}\hat {m}_k^{a\intercal} m_k^{a}+\sum_{a\leq b}\mathbb E_\Delta[\hQ^{ab}_\Delta{Q}^{ab}_\Delta]\\
-\frac{1}{d}\ln\prod_{a=1}^s\int P_w(\bw^a)\dd\bw^a    \left(\prod_{a\leq b}\e^{\mathbb E_\Delta[\Delta\hQ^{ab}_\Delta] \bw^{a\intercal}\bw^b}\prod_{a}\e^{\sqrt d\sum_k{\hM^{a}_k}\bw^{a\intercal}\bmu_k}\right)\\ -\alpha\ln\mathbb E_{k,\Delta}\left[\prod_{a=1}^s\int\dd\eta^a P_k(\eta^a)\mathcal N\left(\bEta\big|m_k^a+b,Q^{ab}_\Delta\right)\right].
\end{multline}
At this point, the free energy $f_\beta$ should be computed functionally extremisizing with respect to all the order parameters by virtue of the Laplace approximation (in addition to $b$),
\begin{equation}
f_\beta=\lim_{s\to 0}\Extr_{\substack{m,\hM\\Q,\hQ,b}}\frac{\Phi^{(s)}(Q,m,\hQ,\hM,b)}{s}.
\end{equation}
However, the convexity of the problem allows us to make an important simplification.
\paragraph{Replica symmetric ansatz} Before taking the $s\to 0$ limit we make the replica symmetric assumptions
\begin{equation}
\begin{split}
Q^{aa}_\Delta&=\begin{cases}R_\Delta,&a=b\\ Q_\Delta&a\neq b\end{cases}\\
m_{k}^{a}&=m_k
\end{split}\qquad
\begin{split}
\hQ^{aa}_\Delta&=\begin{cases}-\frac{1}{2}\hR_\Delta,&a=b\\ \hQ_\Delta&a\neq b\end{cases}\\
\hM_k^{a}&=\hM_k\quad \forall a
\end{split}
\end{equation}
By means of the replica symmetric hypothesis, we can write
\begin{equation}
Q^{ab}_\Delta\mapsto \bsQ_\Delta\equiv(R_\Delta-Q_\Delta)\bI_{s,s}+Q_\Delta\bUno_s.
\end{equation}
The inverse matrix is therefore
\begin{equation}
\bsQ^{-1}_\Delta=  \frac{1}{R_\Delta-Q_\Delta}\bUno_s-\frac{Q_\Delta}{(R_\Delta-Q_\Delta+sQ_\Delta)(R_\Delta-Q_\Delta)}\bI_{s,s},
\end{equation}
whereas
\begin{equation}
\det\bsQ_\Delta=(R_\Delta-Q_\Delta)^{s-1}(R_\Delta-Q_\Delta+sQ_\Delta)
=1+s\ln(R_\Delta-Q_\Delta)+s\frac{Q_\Delta}{R_\Delta-Q_\Delta}+o(s).
\end{equation}
If we denote $V_\Delta\coloneqq R_\Delta-Q_\Delta$
\begin{equation}
\ln\mathbb E_{k,\Delta}\left[\prod_{a=1}^s\int\dd\eta^a P_k(\eta^a)\mathcal N\left(\bEta\big|m_k^a+b,Q^{ab}_\Delta\right)\right]
=s\mathbb E_{k,\Delta,\zeta}\left[\ln Z_k\left(m_k+b+\sqrt{Q_\Delta}\zeta,V_\Delta\right)\right]+o(s),
\end{equation}
with $\zeta\sim\mathcal N(0,1)$ normally distributed random variable. In the expression above, we have also introduced the function
\begin{equation}
Z_k\left(m,V\right)\coloneqq \int\frac{\dd \eta P_k(\eta)}{\sqrt{2\pi V}}\e^{-\frac{(\eta-m)^2}{2V}}.
\end{equation}
On the other hand, denoting by $\hV_\Delta=\hR_\Delta+\hQ_\Delta$,
\begin{multline}\comprimi
    \frac{1}{d}\ln\prod_{a=1}^s\left(\int\dd \bw^a P_w(\bw^a) \e^{-\mathbb E_\Delta[\Delta \hV_\Delta]\frac{ \|\bw^a\|^2}{2}+\sqrt d\sum_k\hM_k\bw^{a\intercal}\bmu_k}\prod_{b}\e^{\frac{1}{2}\mathbb E_\Delta[\Delta \hQ_\Delta]\bw^{a\intercal}\bw^b}\right)=\\
    \comprimi
    =\frac{s}{d}\mathbb E_{\bxi}\ln\left[\int\dd\bw  P_w(\bw)\exp\left(-\mathbb E_\Delta[\Delta \hV_\Delta]\frac{ \|\bw\|^2}{2}+\sqrt d \sum_k\hM_k \bw^\intercal\bmu_k+\sqrt{\mathbb E_\Delta[\Delta \hQ_\Delta]}\bxi^\intercal\bw]\right)\right] +o(s).  
\end{multline}
In the expression above we have introduced $\bxi\sim \mathcal N(\mathbf 0,\bI_d)$ and denote the average over it by $\mathbb E_{\bxi}[\bullet]$. Therefore, the (replicated) \textit{replica symmetric} free-energy is given by
\begin{equation}
\lim_{s\to 0}\frac{\beta}{s}\Phi^{(s)}_{\rm RS}=\sum_{k}\hM_k m_k+\frac{\mathbb E_\Delta\left[\hV_\Delta Q_\Delta-\hQ_\Delta V_\Delta-\hV_\Delta V_\Delta\right]}{2}-\alpha\beta\Psi_{\text{out}}(m,Q,V)-\beta\Psi_w(\hM,\hQ,\hV)
\end{equation}
where we have defined two contributions
\begin{equation}\comprimi\begin{split}
\Psi_{\text{out}}(m,Q,V)&\coloneqq \beta^{-1}\mathbb E_{k,\Delta,\xi}[\ln Z_k\left(\omega_k,V_\Delta\right)]\\
\Psi_w(\hM,\hQ,\hV)&\coloneqq \frac{1}{\beta d}\mathbb E_{\bxi}\ln\left[\int P_w(\bw)\dd\bw \exp\left(-\frac{\mathbb E_\Delta[\Delta \hV_\Delta] \|\bw\|^2}{2}+\sqrt d \sum_k\hM_k \bw^\intercal\bmu_k+\sqrt{\mathbb E_\Delta[\Delta \hQ_\Delta]}\bxi^\intercal\bw\right)\right]
\end{split}\end{equation}
and introduced, for future convenience,
\begin{equation}
\omega_k\coloneqq m_k+b+\sqrt{Q_\Delta}\zeta.
\end{equation}
Note that we have separated the contribution coming from the chosen loss (the so-called \textit{channel} part $\Psi_{\rm out}$) from the contribution depending on the regularisation (the \textit{prior} part $\Psi_w$). To write down the saddle-point equations in the $\beta\to+\infty$ limit, let us first rescale our order parameters as $\hM_k\mapsto \beta\hM_k$, $\hQ_\Delta\mapsto \beta^2\hQ_\Delta$, $\hV\mapsto \beta\hV$ and $V_\Delta\mapsto\beta^{-1}V_\Delta$. For $\beta\to+\infty$ the channel part is
\begin{equation}
\Psi_{\rm out}(m,Q,V)=-\mathbb E_{k,\Delta,\zeta}\left[\frac{(h_k-\omega_k)^2}{2V_\Delta}+\ell_k(h_k)\right].
\end{equation}
where we have written $\Psi_{\rm out}$ in terms of a proximal
\begin{equation}
\arg\min_{u}\left[\frac{(u-\omega_k)^2}{2 V_\Delta}+\ell_k(u)\right].
\label{app:eq:prox}
\end{equation}

A similar expression can be obtained for $\Psi_w$. Introducing the proximal
\begin{equation}
\bgg=\arg\min_{\bw}\left(\frac{\mathbb E_\Delta[\Delta \hV_\Delta] \|\bw\|^2}{2}-\sqrt d \sum_k\hM_k \bw^\intercal\bmu_k-\sqrt{\mathbb E_\Delta[\Delta \hQ_\Delta]}\bxi^\intercal\bw+\lambda r(\bw)\right)\in\R^{d}  
\label{app:eq:proxg}
\end{equation}
We can rewrite the prior contribution $\Psi_w$ as
\begin{equation}
\Psi_w(\hM,\hQ,\hV)
=-\frac{1}{d}\mathbb E_{\bxi}\left[\frac{ \mathbb E_\Delta[\Delta \hV_\Delta]\|\bgg\|^2}{2}-\sqrt d \sum_k\hM_k \bgg^\intercal\bmu_k-\sqrt{\mathbb E_\Delta[\Delta \hQ_\Delta]}\bxi^\intercal\bgg+\lambda r(\bgg)\right]
\end{equation}
The analogy between the two contributions is evident, aside from the different dimensionality of the involved objects. The replica symmetric free energy \eqref{eq:replicaf} in the $\beta\to+\infty$ limit is computed by extremising with respect to the introduced order parameters, 
\begin{equation}\label{eq:app:rsfree}\comprimi
f_{\rm RS}=\Extr_{\substack{m,Q,V,b\\\hM,\hQ,\hV}}\left[\sum_{k=\pm}\hM_k m_k+\frac{\mathbb E_\Delta\left[\hV_\Delta Q_\Delta-\hQ_\Delta V_\Delta\right]}{2}-\alpha\Psi_{\text{out}}(m,Q,V)-\Psi_w(\hM,\hQ,\hV)\right].
\end{equation}
To do so, we have to write down a set of saddle-point equations and solve them.

\paragraph{Saddle-point equations} The saddle-point equations are derived straightforwardly from the obtained free energy functionally extremising with respect to all parameters. A first set of equations is obtained from $\Psi_{\rm out}$. Introducing 
\begin{equation}
f_k\equiv \frac{h_k-\omega_k}{V_\Delta}
\end{equation}
(and keeping in mind that this quantity is also $\Delta$-dependent) we have
\begin{equation}
\hQ_\Delta=\alpha \mathbb E_{k,\zeta}\left[f_k^2\right],\quad
\hV_\Delta=\displaystyle-\frac{\alpha\mathbb E_{k,\zeta}\left[ f_k\zeta\right]}{\sqrt{Q_\Delta}},\quad
\hM_k=\alpha\rho_k\mathbb E_{\Delta,\zeta}\left[f_k\right],\quad
b=\mathbb E_{k,\Delta,\zeta}\left[h_k-m_k\right].
\end{equation}
Denoting $\hat q\coloneqq \mathbb E[\Delta\hQ_\Delta]$ and $\hat v\coloneqq \mathbb E[\Delta\hV_\Delta]$, the saddle-point equations from $\Psi_{w}$ are
\begin{equation}
	V_\Delta =\frac{1}{d}\frac{\Delta\mathbb{E}_{\bxi}[\bgg^\intercal\bxi]}{\sqrt{\hat q}}\qquad
	Q_\Delta =\frac{\Delta}{d} \mathbb{E}_{\bxi}[\|\bgg\|^2]\qquad
	m_k = \frac{1}{\sqrt d}\mathbb{E}_{\bxi}\left[\bgg^\intercal\bmu_k\right].
\end{equation}
Observe that the equations above imply that $V_\Delta=v\Delta $ and $Q_\Delta=q\Delta$, where $v$ and $q$ are some constant that do not depend on $\Delta$\footnote{This was largely expected in our setting, but we preferred to keep a redundant derivation as this factorisation cannot be performed when the derivation is generalised to the case of random covariance matrices which are not multiple of the identity.}. We can rewrite
\begin{equation}\label{app:eq:sp}
	v =\frac{\mathbb{E}_{\bxi}[\bgg^\intercal\bxi]}{d\sqrt{\hat q}}\qquad
	q =\frac{\mathbb{E}_{\bxi}[\|\bgg\|^2]}{d} \qquad
	m_k = \frac{\mathbb{E}_{\bxi}\left[\bgg^\intercal\bmu_k\right]}{\sqrt d}
\end{equation}
and the remaining equations as
\begin{equation}\label{app:eq:sphat}\comprimi
\hat q=\alpha \mathbb E_{k,\zeta,\Delta}\left[\Delta f_k^2\right],\quad
\hat v=\displaystyle-\frac{\alpha}{\sqrt{q}}\mathbb E_{k,\Delta,\zeta}\left[\sqrt\Delta f_k\zeta\right],\quad
\hM_k=\alpha\rho_k\mathbb E_{\Delta,\zeta}\left[f_k\right],\quad
b=\mathbb E_{k,\Delta,\zeta}\left[h_k-m_k\right].    
\end{equation}
To obtain the replica symmetric free energy, therefore, the given set of equation has to be solved, and the result then plugged in Eq.~\eqref{eq:app:rsfree}.

\subsection{Training and test errors}\label{app:replica:errors}
The order parameters introduced to solve the problem allow us to reach our ultimate goal of computing the average errors of the learning process. We will start with the estimation of the training loss
\begin{equation}
\epsilon_\ell\equiv \frac{1}{n}\sum\limits_{\nu=1}^{n}\ell\left(y^{\nu}, \frac{\bw^\star\bx^\nu}{\sqrt d}+b^\star\right)
\end{equation}
in the $n\to+\infty$ limit. The complication in computing this quantity is that the order parameters found in the learning process are, of course, correlated with the dataset $\mathcal D$ used for the learning itself. The best way to proceed is to observe that 
$$
\mathbb E_{\mathcal D}[\mathcal R(\bw^\star,b^\star)]=-\lim_{\beta\to+\infty}\mathbb E_{\mathcal D}[\partial_\beta\ln\mathcal Z_\beta]=\lambda\mathbb E_{\mathcal D}[r(\bw^\star)]+\epsilon_\ell    
$$
where
\begin{equation}
\epsilon_\ell=-\lim_{\beta\to+\infty}\partial_\beta(\beta\Psi_{\rm out})=\lim_{\beta\to+\infty}\mathbb E_{k,\Delta,\zeta}\left[\int \frac{\ell_k(\eta)\e^{-\frac{\beta(\eta-m_k^\star)^2}{2\Delta v^\star}-\beta\ell_k(\eta)}}{\sqrt{2\pi\beta^{-1}v^\star \Delta}\,Z_k(\omega_k^\star,\beta^{-1}v^\star \Delta)}
\dd\eta\right].
\end{equation}
In the $\beta\to+\infty$ limit, the integral concentrates on the minimiser of the exponent, that is, by definition, the proximal $h_k$. In conclusion,
\begin{subequations}
\begin{equation}
\epsilon_\ell=\mathbb E_{k,\Delta,\bxi}[\ell_k(h_k)].    
\end{equation}
By means of the same concentration result, the training error is
\begin{equation}\label{app:eq:et}
\epsilon_t=\frac{1}{n}\sum_{\nu=1}^n\mathbb I\left(\varphi\left(\frac{\bw^{\star\intercal}\bx^\nu}{\sqrt d}+b^\star\right)\neq y^\nu\right)\xrightarrow{n\to+\infty}\mathbb E_{k,\Delta,\zeta}\left[\mathbb I(\varphi(h_k)\neq y_k)\right].
\end{equation}
The expressions above hold in general, but, as anticipated, important simplifications can occur in the set of saddle-point equations \eqref{app:eq:sphat} and \eqref{app:eq:sp} depending on the choice of the loss $\ell$ and of the regularisation function $r$. The generalisation (or test) error can be written instead as
\begin{equation}
\epsilon_g=\mathbb E_{(y^{\rm new},\bx^{\rm new})}\left[\mathbb I\left(\varphi\left(\frac{\bw^{\star\intercal}\bx^{\rm new}}{\sqrt d}+b^\star\right)\neq y^{\rm new}\right)\right].
\end{equation}
This expression can be rewritten as
\begin{equation}\begin{split}
\epsilon_g&=\mathbb E_{k}\left[\int\mathbb I(\varphi(\eta)=y_k)\mathbb E_{\bx^{\rm new}}\left[\delta\left(\eta-\frac{\bw^{\star\intercal}\bx^{\rm new}}{\sqrt d}-b^\star\right)\right]\dd\eta\right]\\
&=\mathbb E_{k,\Delta}\left[\int\mathbb I(\varphi(\eta)=y_k)\mathbb E_{\bx^{\rm new}|\Delta}\left[\delta\left(\eta-\frac{\bw^{\star\intercal}\bx^{\rm new}}{\sqrt d}-b^\star\right)\right]\dd\eta\right]\\
\xrightarrow{d\to+\infty}&\phantom{=}\mathbb E_{k,\Delta}\left[\int\mathbb I(\varphi(\eta)=y_k)\mathcal N(\eta|m^\star_k+b^\star,\Delta q^\star)\dd\eta\right]\\
&=\mathbb E_{k,\Delta,\zeta}\left[\mathbb I\left(\varphi\left(m_k^\star+\sqrt{\Delta q^\star}\zeta+b^\star\right)\neq y_k\right)\right].
\end{split}
\end{equation}
\label{app:eq:errori}
\end{subequations}
This can be easily computed numerically once the order parameters (including their functional dependence on $\Delta$) are given.

\subsection{Ridge regularisation}\label{app:ridge_regul}
\begin{wrapfigure}{r}{0.5\textwidth}
    \vspace{-2cm}\centering
    \includegraphics[width=0.49\textwidth]{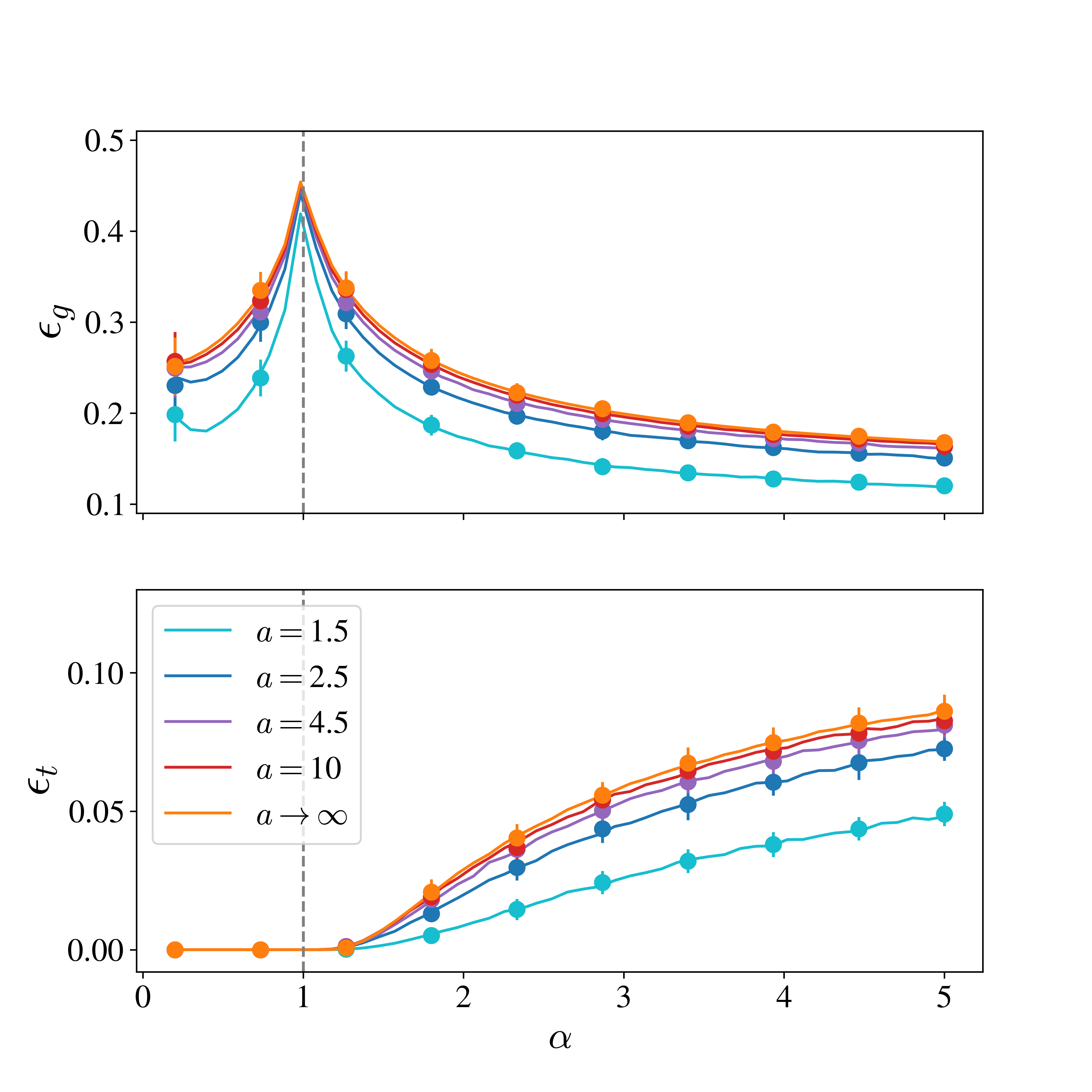}
    \caption{Test error $\epsilon_g$ (\textit{top}) and training error $\epsilon_t$ (\textit{bottom}) as predicted by Eq.~\eqref{eq:errori} in the unbalanced $\rho=\nicefrac{1}{4}$ case. The dataset distribution is parametrised as in Eq.~\eqref{eq:invgamma0}. The classification task is solved using a quadratic loss with ridge regularisation with $\lambda=10^{-5}$. Dots correspond to the average outcome of $50$ numerical experiments in dimension $d=10^3$. In our parametrisation, the population covariance is $\bSigma=\bI_d$ for all values of $a$ and moreover, for $a\to+\infty$, the case of pure Gaussian clouds $P(\bx|\bmu)=\mathcal N(\bx|\bmu,\bI_d)$ is recovered.}\label{app:fig:ridge}
    \vspace{-2.5cm}
\end{wrapfigure}
{Let us fix now $r(\bw)=\frac{1}{2}\|\bw\|^2$. In this case, the computation of $\Psi_w$ can be performed explicitly via a Gaussian integration, and the saddle-point equations can take a more compact form that is particularly suitable for a numerical solution. In particular
\begin{equation}
\bgg
= \frac{\sqrt d\sum_k \hM_k\bmu_k+\sqrt{\hat q}\bxi}{\lambda+\hat v}
\end{equation}
so that the prior saddle-point equations obtained from $\Psi_w$ become
\begin{subequations}\label{app:eq:sp_l2}
\begin{align}
	v &=\frac{1}{\lambda+\hat v}\\
	q &=\frac{\sum_{kk'}\hM_{k}\hM_{k'}\bmu_k^\intercal\bmu_{k'}+\hat q}{(\lambda+\hat v)^2}\\
	m_k &= \frac{\sum_{k'}\hM_{k'}\bmu_{k'}^\intercal\bmu_k}{\lambda+\hat v}.
\end{align}
\end{subequations}

\paragraph{Quadratic loss}
If we consider a quadratic loss $\ell(y,x)=\frac{1}{2}\left(y-x\right)^2$, then an explicit formula for the proximal can be found, namely
\begin{equation}
f_k=\frac{y_k-\omega_k}{1+v\Delta}
\end{equation}
so that the second set of saddle-point equations \eqref{app:eq:sphat} can be written as
\begin{subequations}
\begin{align}
\hat q&\comprimi=\alpha \mathbb E_{k,\Delta}\left[\frac{\Delta(y_k-m_k-b)^2}{(1+v\Delta)^2}\right]+\alpha q\mathbb E_\Delta\left[\frac{\Delta^2}{(1+v\Delta)^2}\right], 
\\
\hat v&=\alpha\mathbb E_\Delta\left[\frac{\Delta}{1+v\Delta }\right]=\frac{1-v\lambda}{v},\\
\hM_k&=\alpha\rho_k(y_k-m_k-b)\mathbb E_{\Delta}\left[\frac{1}{1+v\Delta}\right].
\end{align}
\end{subequations}
so that $v$ satisfies the self-consistent equation
\begin{equation}
\frac{1-v\lambda}{v\alpha}=\mathbb E_\Delta\left[\frac{\Delta}{1+v\Delta}\right].   
\end{equation}
As a complement to the information given in the main text, in Fig.~\ref{app:fig:ridge} we give some numerical results for the case of \textit{unbalanced} clusters, which show a perfect agreement between the theoretical predictions and the numeral results.

\paragraph{Logistic loss} \label{app:logistic} Let us now consider $K=2$ and $\mathcal Y=\{-1,1\}$: in the following, we will label the different classes by $k=\pm$. In this context, we discuss the relevant case of the logistic loss $\ell_\pm(x) = \ln(1+\e^{\mp x})$.
The proximal equation for this loss is the solution of the equations:
\begin{equation}
f_\pm=\arg\min_{u}\left[\frac{\Delta v u^2}{2}+\ln(1+\e^{\mp (\Delta vu+\omega_\pm)})\right]
\end{equation}
having only one solution for which, however, there is no closed-form expression; the equation can be solved numerically. Interestingly, the $\lambda\to 0$ limit of such loss recovers the hinge loss with zero margin \citep{rosset2003margin}. It is numerically convenient in this case to consider $\hat v \mapsto \lambda \hat v$, $v \mapsto \lambda^{-1} v$, $m \mapsto \lambda^{-1} m$, $b\to\lambda^{-1}b$ and $q \mapsto \lambda^{-2} q$: this provides a new set of consistent equations for the rescaled variables, implying that in the $\lambda\to 0$ limit the original order parameters diverge.

The zero-regularisation limit of the logistic loss can help us study the separability transition. To obtain the position of the separability transition, we follow the derivation proposed by \citet{mignacco20a}. Let us assume for simplicity that $\bmu_+=-\bmu_-\equiv \bmu$ with $\|\bmu\|=1$ for simplicity. It is immediate to see that in this case $m_+=-m_-\equiv m$, and it is, therefore, possible to introduce $\hat m\coloneqq \hat m_++\hat m_-$, so that the saddle-point equations can be written as
\begin{equation}
\begin{cases}
\hat q=\alpha \mathbb E_{\pm,\zeta,\Delta}\left[\Delta f_\pm^2\right],\\
\hat v=\displaystyle-\frac{\alpha}{\sqrt{q}}\mathbb E_{\pm,\Delta,\zeta}\left[\sqrt\Delta f_\pm\zeta\right],\\
\hat m=\alpha\mathbb E_{\pm,\Delta,\zeta}\left[f_\pm\right],\\
b=\mathbb E_{\pm,\Delta,\zeta}\left[h_\pm\mp m\right],
\end{cases}\qquad 
\begin{cases}
	v =\frac{1}{\lambda+\hat v}\\
	q =\frac{\hat m^2+\hat q}{(\lambda+\hat v)^2}\\
	m= \frac{\hat m}{\lambda+\hat v}.
\end{cases}  
\end{equation}
We can start with the fact that
\begin{equation}
\alpha v^2\mathbb E_{\pm,\zeta,\Delta}[\Delta f_\pm^2]=v^2\hat q=q-m^2.
\end{equation}
Introducing $\theta=\frac{m}{\sqrt q}$, $\tilde b=\frac{b}{\sqrt q}$ and $\tilde v=\frac{v}{\sqrt q}$, then we can re-write the equation as
\begin{equation}
\alpha\mathcal S(\tilde v,\theta,\tilde b,q)=1-\theta^2,\qquad\mathcal S(\tilde v,\theta,\tilde b,q)\coloneqq\tilde v^2 \mathbb E_{\pm,\zeta,\Delta}[\Delta f_\pm^2].\label{app:eqv}
\end{equation} 

We state now that the separable phase corresponds to the limit $\tilde v\to+\infty$ as $\lambda\to 0$ when using the logistic loss, keeping $\theta$ and $b$ fixed. In this case, indeed, we have that $f_\pm=\ell'_\pm(h_\pm)\to 0$, i.e., $h_\pm\to\pm+\infty$, which is the condition to have separability, $\lim_{\lambda\to 0}\epsilon_\ell=\lim_{\lambda\to 0}\mathbb E[\ell_\pm(h_\pm)]=0$. 

To properly take this limit we need an explicit expression of $\mathcal S(\theta,\tilde b,q)$, we observe that $h_\pm$ satisfies $h_\pm+ v\Delta \ell'_\pm(h_\pm)=\omega_\pm$, with $\omega_\pm\coloneqq \pm m+b+\sqrt{q\Delta}\zeta$ and $f_\pm\coloneqq \ell_\pm'(h_\pm)$: it is possible therefore introduce the Legendre function $\tilde \ell_\pm(f)=\max_h\{hf-\ell_\pm(h)\}$ of $\ell_\pm(h)$, such that $v\Delta\,f_\pm+\tilde \ell'_\pm(f_\pm)=\omega_\pm$. If now we want to compute the probability that $f_\pm<f$ at fixed label and value of $\Delta$, we have
\begin{equation}
\begin{split}
\mathbb P[f_\pm\leq f|\pm,\Delta]
&=\mathbb P[\pm m+b+\sqrt{q\Delta}\zeta\leq v\Delta\,f+\tilde\ell'(f)|\pm,\Delta]\\  
&=\Phi\left(\frac{v\Delta f\mp m-b+\tilde\ell'_\pm(f)}{\sqrt{q\Delta}}\right),
\end{split}\end{equation}
with $\Phi(z)=\mathbb P[\zeta\leq z]$ for a standard Gaussian random variable $z\sim\mathcal N(0,1)$. By consequence, using the fact that $\mathbb E_\zeta[f^2_\pm]=-2\int_{-1}^0 f\mathbb P[f_\pm \leq f|\pm,\Delta]\dd f$, and observing that $-1\leq f_\pm\leq 0$
\begin{equation}\comprimi
\begin{split}
\mathcal S(\tilde v,\theta,\tilde b,q)&\coloneqq2\tilde v^2\int_0^{1}f\mathbb E_\Delta\left[\Delta\rho_+\Phi\left(\frac{-\tilde v\Delta f-\theta-\tilde b}{\sqrt{\Delta}}+\frac{\tilde\ell'_+(-f)}{\sqrt{2q\Delta}}\right)+\Delta\rho_-\Phi\left(\frac{-\Delta\tilde v f+\theta-\tilde b}{\sqrt{\Delta}}+\frac{\tilde\ell'_-(-f)}{\sqrt{2q\Delta}}\right)\right]\dd\\
&=2\int_0^{\tilde v}f\mathbb E_\Delta\left[\Delta\rho_+\Phi\left(\frac{-\Delta f-\theta-\tilde b}{\sqrt{\Delta}}+\frac{\tilde\ell'_+(-\sfrac{f}{\tilde v})}{\sqrt{2q\Delta}}\right)+\Delta\rho_-\Phi\left(\frac{-\Delta f+\theta-\tilde b}{\sqrt{\Delta}}+\frac{\tilde\ell'_-(-\sfrac{f}{\tilde v})}{\sqrt{2q\Delta}}\right)\right]\dd f.
\end{split}\end{equation}
Eq.~\eqref{app:eqv} allows us to express $\tilde v\equiv \tilde v(\theta,b,q)$. It turns out that $\tilde v=\tilde v(\theta,\tilde b,q)$ is finite for any finite $q$, and, at $\tilde b$ and $\theta$ fixed, can diverge for $q\to+\infty$ only. If we assume $q$ to be finite and we take $\lim_{\tilde v\to+\infty}\mathcal S(\tilde v,\theta,\tilde b,q)$, the expression diverges giving therefore an inconsistency (as $\tilde \ell'_\pm(-\sfrac{f}{\tilde v})\to\pm \infty$). On the other hand, by taking the limit $q\to+\infty$, the function $\lim_{q\to+\infty}\mathcal S(\tilde v,\theta,\tilde b,q)$ has a finite limit and it is monotonically increasing in $\tilde v$, so that Eq.~\eqref{app:eqv} allows for a finite $\tilde v$ as long as
\begin{equation}
\alpha>\frac{1-\theta^2}{\mathcal S_\star(\theta,\tilde b)}, 
\end{equation}
where
\begin{equation}\comprimi\begin{split}
\mathcal S_\star(\theta,\tilde b)&=\lim_{\tilde v\to+\infty}\lim_{q\to+\infty}\mathcal S(\tilde v,\theta,\tilde b,q)\\
&=2\int_0^\infty f\mathbb E_\Delta\left[\Delta\rho_+\Phi\left(\frac{-\Delta f-\theta-\tilde b}{\sqrt{\Delta}}\right)+\Delta\rho_-\Phi\left(\frac{-\Delta f+\theta-\tilde b}{\sqrt{\Delta}}\right)\right]\dd f\\
&=\int_0^\infty f^2\mathbb E_\Delta\left[\rho_+\mathcal N\left(f+\frac{\theta+\tilde b}{\sqrt{\Delta}}\Big|0,1\right)+\rho_-\mathcal N\left(f+\frac{\theta-\tilde b}{\sqrt{\Delta}}\Big|0,1\right)\right]\dd f.
\end{split}\end{equation}
As a result, given that $\theta\in(0,1]$, the smaller value for which $\tilde v$ is finite is
\begin{equation}
\alpha^\star=\max_{\theta\in(0,1],b}\frac{1-\theta^2}{\mathcal S_\star(\theta,\tilde b)}.    
\end{equation}
This corresponds to the threshold value for the separability transition.
\subsection{The Bayes optimal error in the $K=2$ case}\label{app:sec:bayes}
We derive here the Bayes optimal error in the case of two clusters $K=2$ with centroids in $\sfrac{1}{\sqrt d}\bmu_\pm=\pm\sfrac{1}{\sqrt{d}}\bmu$, with $\bmu\sim\mathcal N(\mathbf 0,\bI_d)$. The derivation is a variation of the arguments in Ref.~\cite{mignacco20a}.  Given an estimate of $\bmu$, it is possible to compute $p(y,\bx|\bmu)=p(\bx|y,\bmu)p(y)=\mathbb E[\mathcal N(\bx;\sfrac{y}{\sqrt d}\bmu,\Delta\bI)]p(y)$ with $p(y)=\rho\delta_{y,+1}+(1-\rho)\delta_{y,-1}$. We assume for now that both $\mathbb E[\Delta]$ and $\mathbb E[\Delta^{-2}]$ are finite. We can write down the posterior for $\bmu$ given a dataset $\mathcal D=\{(y_\nu,\bx_\nu)\}_{\nu=1}^n$ of observations as
\begin{equation}
p(\bmu|\mathcal D)\propto p(\mathcal D|\bmu)p(\bmu)\propto \mathbb E\left[\frac{\exp\left(-\frac{\|\bmu\|^2}{2}-\sum\limits_{\nu=0}^n\frac{\|\bx_\nu-y_\nu\bmu\|^2}{2\Delta_\nu}\right)}{\prod_\nu(2\pi\Delta_\nu)^{d/2}}\right]
\end{equation}
where the expectation is over the set $\bDelta\coloneqq\{\Delta_\nu\}_{\nu\in[n]}$. Given a new pair $(y_0,\bx_0)$, we can now estimate
\begin{equation}
\mathbb P[y_0=\pm 1|(\bx_0,\Delta_0),\{(y_\nu,\bx_\nu,\Delta_\nu\}_{\nu=1}^n]\propto 
p(y_0)\int\dd\bmu\frac{\e^{-\frac{\|\bmu\|^2}{2}-\sum\limits_{\nu=0}^n\frac{1}{2\Delta_\nu}\left\|\bx_\nu-\frac{y_\nu\bmu}{\sqrt d}\right\|^2}}{\prod_\nu(2\pi\Delta_\nu)^{d/2}},
\end{equation}
where we condition on the values of the set $\{\Delta_\nu\}_{\nu=0}^n$ as well. The integral can be computed as
\begin{equation}
\int\dd\bmu\e^{-\frac{\|\bmu\|^2}{2}-\sum_{\nu=0}^n\frac{1}{2\Delta^\nu}\left\|\bx_\nu-\frac{y_\nu\bmu}{\sqrt d}\right\|^2}\propto \frac{1}{(1+\nicefrac{\alpha}{\delta_n})^{d/2}}\exp\left(\frac{\alpha y_0\bx_0}{\sfrac{1}{d}+\Delta_0(1+\nicefrac{\alpha}{\delta_{n}})}\,\frac{1}{n}\sum_{\nu=1}^n\frac{y_\nu\bx_\nu}{\Delta_\nu}\right)
\end{equation}
where we have omitted a coefficient independent of $y_0$, and we have introduced the harmonic mean
\begin{equation}
\delta_n\coloneqq\left(\frac{1}{n}\sum_{\nu=1}^n\frac{1}{\Delta_\nu}\right)^{-1}.    
\end{equation}
For the sake of brevity, let us an auxiliary random variable correlated with the harmonic mean, namely
$$\hat\delta_n\coloneqq\left(\frac{1}{n}\sum_{\nu=1}^n\frac{\Delta_\nu'}{\Delta_\nu^2}\right)^{-1}$$ 
where $\{\Delta'_\nu\}_{\nu\in[n]}$ is a distinct and independent set of values of $\Delta$'s. Now we have that, by writing for $\nu\in[n]$ $\bx_\nu=y_\nu\bmu+\sqrt{\Delta_\nu'}\bzeta_\nu$, $\bzeta_\nu$ vector distributed as $\mathcal N(\mathbf 0,\bI_d)$, and $\bx_0=y_0^\star\bmu+\sqrt{\Delta_0^\star}\bzeta_0$, $\bzeta_0$ also distributed as $\mathcal N(\mathbf 0,\bI_d)$, then up to $O(\sfrac{1}{d})$ contributions
\begin{equation}
\begin{split}
\frac{1}{n}\sum_{\nu=1}^n\frac{y_\nu \bx_0^\intercal\bx_\nu}{\Delta_\nu}&=\frac{y_0^\star}{\delta_n}+\frac{y_0^\star}{n}\sum_{\nu=1}^n\frac{y_\nu \sqrt{\Delta_\nu'}\bmu^\intercal\bzeta_\nu}{\Delta_\nu}+\frac{\sqrt{\Delta_0^\star}\bmu^\intercal\bzeta_0}{\delta_n}+\frac{1}{n}\sum_{\nu=1}^n\frac{y_\nu\sqrt{\Delta_\nu'\Delta_0^\star}\bzeta_0^\intercal\bzeta_\nu}{\Delta_\nu}+O(\nicefrac{1}{d})\\
&\stackrel{\rm d}{=}\frac{y_0^\star}{\delta_n}+\sqrt{\frac{\Delta_0^\star}{\delta_n^2}+\frac{\Delta_0^\star}{\alpha \hat\delta_n}}\zeta+O(\nicefrac{1}{d})=\frac{y_0^\star+\sqrt{\Delta_0^\star B(\delta_n,\hat\delta_n)}}{\delta_n}+O(\nicefrac{1}{d}),
\end{split}
\end{equation}
at the leading order in $n,d$, where $\zeta\sim\mathcal N(0,1)$. We have also introduced $B(\delta,\hat\delta)\coloneqq 1 + \frac{\delta^{2}}{\alpha \hat\delta}$. In the large $n,d$ limit therefore
\begin{equation}\comprimi 
\mathbb P[y_0=\pm 1|(\bx_0,\Delta_0),\{(y_\nu,\bx_\nu,\Delta_\nu\}_{\nu=1}^n]\propto 
\exp\left[\frac{y_0 A(\delta_n) }{\Delta_0}\left(y_0^\star+\sqrt{\Delta_0^\star B(\delta_n,\hat\delta_n)}\zeta\right)+\ln p(y_0)\right]
\end{equation}
where $A(\delta)\coloneqq\frac{\alpha}{\alpha+\delta}$. The conditional optimal estimator is then 
\begin{equation}\hat y_0|_{\Delta_0,\bDelta}=\arg\max_{y_0\in\{-1,1\}}\left[\frac{y_0 A(\delta_n)}{\Delta_0}\left(y_0^\star+\sqrt{\Delta_0^\star B(\delta_n,\hat\delta_n)}\zeta\right)+\ln p(y_0)\right]
\end{equation}
the dependence on $\bx^0$ being expressed by $\zeta$ and $\Delta_0^\star$. The probability that such an estimator is in fact not exact is
\begin{equation}
\mathbb P[\hat y_0\neq y_0^\star|\Delta_0,\bDelta]=\mathbb P\left[y_0^\star\zeta<-\frac{1}{\sqrt{\Delta_0^\star B(\delta_n,\hat\delta_n)}}\left(1+\frac{\Delta_0
}{2A(\delta_n) }\ln\frac{p(y^\star_0)}{p(-y_0^\star)}\right)\right].  
\end{equation}
If $\hat\Phi(x)\coloneqq1-\Phi(x)=\frac{1}{\sqrt{2\pi}}\int_x^\infty\e^{-\sfrac{t^2}{2}}\dd t$, then the Bayes optimal error is therefore
\begin{equation}
\varepsilon_{\rm g}^{\rm BO}=\rho\mathbb E\left[\hat\Phi\left(\kappa_+\right)\right]+(1-\rho)\mathbb E\left[\hat\Phi\left(\kappa_-\right)\right],  
\end{equation}
where, observing that $\delta_n^{-1}\to\mathbb E[\Delta^{-1}]$ and $\hat\delta_n^{-1}\to \mathbb E[\Delta]\mathbb E[\Delta^{-2}]$,
$$\kappa_\pm\equiv \kappa_\pm(\Delta_0,\Delta_0^\star)\coloneqq\frac{1\pm \frac{\Delta_0}{2}\left(1+\frac{1}{\alpha\mathbb E[\Delta^{-1}]}\right)\ln\frac{\rho}{1-\rho}}{\sqrt{\Delta_0^\star \left(1+\frac{\mathbb E[\Delta]\mathbb E[\Delta^{-2}]}{\alpha\mathbb E[\Delta^{-1}]^2}\right)}}.$$

\subsection{The uncorrelated-teacher case and universality in binary classification}\label{App:RL}
In the present section, we will focus on the $K=2$ case, assuming labels to be given by $\mathcal Y=\{-1,+1\}$. In a recent paper, \citet{gerace2023} showed that a classification task on Gaussian clouds exhibits universality features in the case in which the labels are randomly assigned to the dataset points. Under the hypothesis of a loss function satisfying $\ell(y,
\eta)=\ell(-y,-\eta)$ with ridge regularisation, they show that a dataset obtained from a mixture of Gaussian clouds with equal covariance $\bSigma$ is equivalent to a dataset obtained from a single Gaussian cloud with zero mean and covariance $\bSigma$. Moreover, using ridge regression, the training loss $\epsilon_\ell$ is shown to depend on the sample complexity only (and not on $\bSigma$) in the $\lambda\to 0^+$ limit. This result has been generalised immediately afterwards by \citet{pesce2023}, who showed that the same picture holds in the case in which labels are generated by a ``teacher'' modeled by a distribution $P_0(y|\tau)$, with $P_0(y|\tau)=P_0(-y|-\tau)$, and parametrised by vector $\btheta_0\in\R^d$, which is \textit{uncorrelated} with the data structure. 

In our setting, this would amount to considering a database $\mathcal D$ generated from the joint distribution
\begin{equation}
P(\bx,y)=P_0\left(y\Big|\frac{\btheta_0^\intercal\bx}{\sqrt d}\right)\sum_{k=\pm}\rho_k\mathbb E[\mathcal{N}\left(\bx\left|\bmu_k,\Delta\bI_d\right.\right)],  
\end{equation}
the random labels case corresponding to $P_0(y|\tau)$ given by a Rademacher distribution in $y$ independent of $\tau$. The condition of uncorrelated teacher is expressed by\footnote{In our case, this condition is simpler than in Ref.~\citep{pesce2023} as we assume that each class has the same homogeneous covariance.}
\begin{equation}
\lim_{d\to+\infty}\frac{\btheta_0^\intercal\bmu_\pm}{d}=0.    
\end{equation}
Let us assume for simplicity that we adopt ridge regularisation, $r(\bw)=\frac{\lambda}{2}\|\bw\|^2$. The analysis of this setting is perfectly analogous to the one discussed above, but provides slightly different fixed-point equations for the order parameters, and in particular, it requires introducing the overlap between the weights $\bw$ and the teacher parameter $\btheta_0$, $t=\frac{1}{d}\bw^\intercal\btheta_0$, and its corresponding Lagrange multiplier $\hat t$. Following therefore a procedure that combines the one presented above and the derivation given in Ref.~\citep{pesce2023} for the Gaussian case, we can obtain the following fixed-point equations,
\begin{equation}\comprimi
\begin{cases}
v =\frac{1}{\lambda+\hat v}\\
q =\frac{\gamma\hat t^2+\|\sum_{c}\hat m_c\bmu_{c}\|^2+q}{(\lambda+\hat v)^2},\\
m_\pm = \frac{\sum_{c'}\hat m_{c'}\bmu_{c'}^\intercal\bmu_\pm}{\lambda+\hat v},\\
t = \frac{\gamma\hat t}{\lambda+\hat v}
\end{cases} \quad
\begin{cases}
\hat q=\alpha  \sum_y\mathbb E_{\pm,\Delta ,\zeta}\left[\Delta Z_0\left(y,\sqrt{\frac{\Delta}{q}}t\zeta,\Delta \gamma-\Delta \frac{t^2}{q}\right)f^2(y,\omega_\pm,v)\right],\\
\hat v=-\alpha  \sum_y\mathbb E_{\pm,\Delta ,\zeta}\left[Z_0\left(y,\sqrt{\frac{\Delta }{q}}t\zeta,\Delta \gamma-\Delta \frac{t^2}{q}\right)\partial_\omega f(y,\omega_\pm,v)\right],\\
\hat m_\pm=\alpha  \rho_\pm\sum_y\mathbb E_{\Delta ,\zeta}\left[Z_0\left(y,\sqrt{\frac{\Delta }{q}}t\zeta,\Delta \gamma-\Delta \frac{t^2}{q}\right)f_\pm(y,\omega_\pm,v)\right]\\
\hat t=\alpha\sum_y\mathbb E_{\pm,\Delta ,\zeta}\left[\partial_\omega Z_0\left(y,\sqrt{\frac{\Delta }{q}}t\zeta,\Delta \gamma-\Delta \frac{t^2}{q}\right)\Delta  f(y,\omega_\pm,v)\right],
\end{cases}
\end{equation}
with $\gamma\coloneqq\frac{1}{d}\|\btheta_0\|^2$ and
\begin{equation}
\begin{split}
f(y,\omega,v)&\textstyle\comprimi\coloneqq \arg\min_{u}\left[\frac{\Delta v u^2}{2}+\ell(y,\omega+v\Delta u)\right],\quad \omega_\pm\coloneqq b+m_\pm+\sqrt{\Delta q}\zeta\\
b&\textstyle\comprimi=\sum_y\mathbb E_{\pm,\Delta,\zeta}\left[Z_0\left(y,\sqrt{\frac{\Delta }{q}}t\zeta,\Delta \rho-\Delta \frac{t^2}{q}\right)\left(\omega_\pm+v\Delta f(y,\omega_{\pm},v)-m_\pm\right)\right].
\end{split}
\end{equation}
that provide us the quantities to plug into Eqs.~\eqref{app:eq:errori} to compute the asymptotic errors. In the equations above, $Z_0(y,\omega,v)\coloneqq \mathbb E_z[P_0(y|\omega+\sqrt{v}z)]$ with $z\sim\mathcal N(0,1)$. 

\paragraph{Mean universality} Following Ref.~\citep{pesce2023}, let us now make the additional assumption $P_0(y|\tau)=P_0(-y|-\tau)$. This implies that $Z_0(y,\omega,v)=Z_0(-y,-\omega,v)$. In this case, we claim that the solution has $m_\pm=\hat m_\pm=b=0$. It is clear that if $\hat m_\pm=0$, then $m_\pm =0$ and, moreover, because of parity, the solution $b=0$ is consistent. On the other hand, if $m_\pm=0$ and $b=0$, remembering that $\ell(y,\eta)=\ell(-y,-\eta)$, then $f(1,\omega,v)=-f(-1,-\omega,v)$ so that the formula for $\hat m_\pm$ involves a Gaussian integral of an odd function which is, therefore, zero, implying $\hat m_\pm=0$. We are left therefore with a much simpler set of fixed-point equations, namely
\begin{equation}\comprimi
\begin{cases}
v =\frac{1}{\lambda+\hat v}\\
q =\frac{\gamma\hat t^2+q}{(\lambda+\hat v)^2},\\
t = \frac{\gamma\hat t}{\lambda+\hat v}
\end{cases} \quad
\begin{cases}
\hat q=\alpha  \sum_y\mathbb E_{\Delta ,\zeta}\left[\Delta Z_0\left(y,\sqrt{\frac{\Delta}{q}}t\zeta,\Delta \gamma-\Delta \frac{t^2}{q}\right)f^2(y,\omega,v)\right],\\
\hat v=-\alpha  \sum_y\mathbb E_{\Delta ,\zeta}\left[Z_0\left(y,\sqrt{\frac{\Delta }{q}}t\zeta,\Delta \gamma-\Delta \frac{t^2}{q}\right)\partial_\omega f(y,\omega,v)\right],\\
\hat t=\alpha\sum_y\mathbb E_{\Delta ,\zeta}\left[\partial_\omega Z_0\left(y,\sqrt{\frac{\Delta }{q}}t\zeta,\Delta \gamma-\Delta \frac{t^2}{q}\right)\Delta  f(y,\omega,v)\right],
\end{cases}
\end{equation}
with
\begin{equation}\textstyle\comprimi
f(y,\omega,v)\coloneqq \arg\min_{u}\left[\frac{\Delta v u^2}{2}+\ell(y,\omega+v\Delta u)\right],\quad \omega\coloneqq \sqrt{\Delta q}\zeta,
\end{equation}
which is exactly the formula we would have obtained assuming $\bmu_\pm=\mathbf 0$, i.e., the presence of \textit{one cloud only}, so that $P(\bx,y)=P_0\big(y\big|\frac{1}{\sqrt d}\btheta_0^\intercal\bx\big)\mathbb E[\mathcal{N}\left(\bx\left|\mathbf 0,\Delta\bI_d\right.\right)]  $. This \textit{mean universality} result generalises the result in Ref.~\citep{pesce2023} for Gaussian clouds.

\paragraph{Random labels under square loss} The case of \textit{random label} is particularly interesting as it exhibits further universality when the square loss is adopted. In this case, $P_0(y|\tau)$ is simply the Rademacher distribution. Using $\ell(y,\eta)=\frac{1}{2}(y-\eta)^2$, the equations greatly simplify and we obtain
\begin{equation}\textstyle\comprimi
\begin{cases}
v =\frac{1}{\lambda+\hat v},\\
q =\frac{\hat q}{(\lambda+\hat v)^2},\\
\end{cases} \ 
\begin{cases}
\hat q=\alpha \mathbb E_\Delta\left[\frac{\Delta+q\Delta^2}{(1+v\Delta)^2}\right],\\
\hat v=\alpha\mathbb E_\Delta\left[\frac{\Delta}{1+v\Delta }\right],
\end{cases}
\end{equation}
whereas $t=\hat t=0$, so that the test loss, obtained by using $\varphi(y,\eta)=(y-\eta)^2$, and the training loss are
\begin{equation}
\hat\epsilon_g\coloneqq\mathbb E_{(y,\bx)}\left[\left(y-\frac{1}{\sqrt d}\bx^\intercal\bw^\star\right)\right]=1+\sigma^2 q,\qquad\epsilon_\ell=\frac{1}{2}\mathbb E_\Delta
\left[\frac{1+q\Delta }{(1+v\Delta)^2}\right].
\end{equation}
Note that the test loss is infinite if $\sigma^2=+\infty$. Introducing the notation $\delta_k\coloneqq\mathbb E[(1+v\Delta)^{-k}]$, the fixed-point equations above read
\begin{equation}\textstyle\comprimi
\begin{cases}
v =\frac{1}{\lambda+\hat v},\\
q =\frac{\hat q}{(\lambda+\hat v)^2},\\
\end{cases} \ 
\begin{cases}
\hat q=\alpha\frac{\delta_1-\delta_2}{v}+\alpha q\frac{1-2\delta_1+\delta_2}{v^2},\\
\hat v=\alpha\frac{1-\delta_1}{v},
\end{cases},\quad \epsilon_\ell=\frac{1}{2}\left(\delta_2+q\frac{\delta_1-\delta_2}{v}\right).
\end{equation}
Observe now that in the limit $\lambda\to 0$
\begin{equation}
x\coloneqq \frac{q}{v}=\frac{\hat q}{\hat v}=-\frac{\delta_2-2\delta_1+1}{\delta_1-1}x+\frac{\delta_2-\delta_1}{\delta_1-1},
\end{equation}
which is solved by $x=1$, so that in this limit $\epsilon_\ell=\frac{1}{2}\delta_1$. But, on the other hand, the fixed point equation for $\hat v$ implies that $v$ is such that $\delta_1=1-\sfrac{1}{\alpha}$ if $\alpha\geq 1$, and zero otherwise (as $\delta_1\geq 0$ by definition) so that we recover the universal formula for the training loss obtained by \citet{gerace2023},
\begin{equation}\label{app:loss:gera}
\epsilon_\ell=\frac{1}{2}\left(1-\frac{1}{\alpha}\right)_+,    
\end{equation}
where $(x)_+=x\theta(x)$. Note that this formula \textit{does not depend on the choice of the distribution of $\Delta$}. We verify this result in Fig.~\ref{app:fig:rl}. We run numerical experiments using a quadratic loss on datapoints split in two clouds of equal weights centered around $\bmu_1=-\bmu_2\sim\mathcal N(\mathbf 0,\bI_d)$, and generated with distribution as in \eqref{eq:invgamma0}, parametrised by $a$, so that each cloud has $\bSigma=\bI_d$. Labels are assigned randomly with Rademacher distribution. The results are compared with the prediction for \textit{one} cloud with $\bmu=\mathbf 0$ but with the same parameter $a$. We see that mean-independence in this setting is indeed verified.

\begin{figure}
    \centering
    \includegraphics[width=0.8\textwidth]{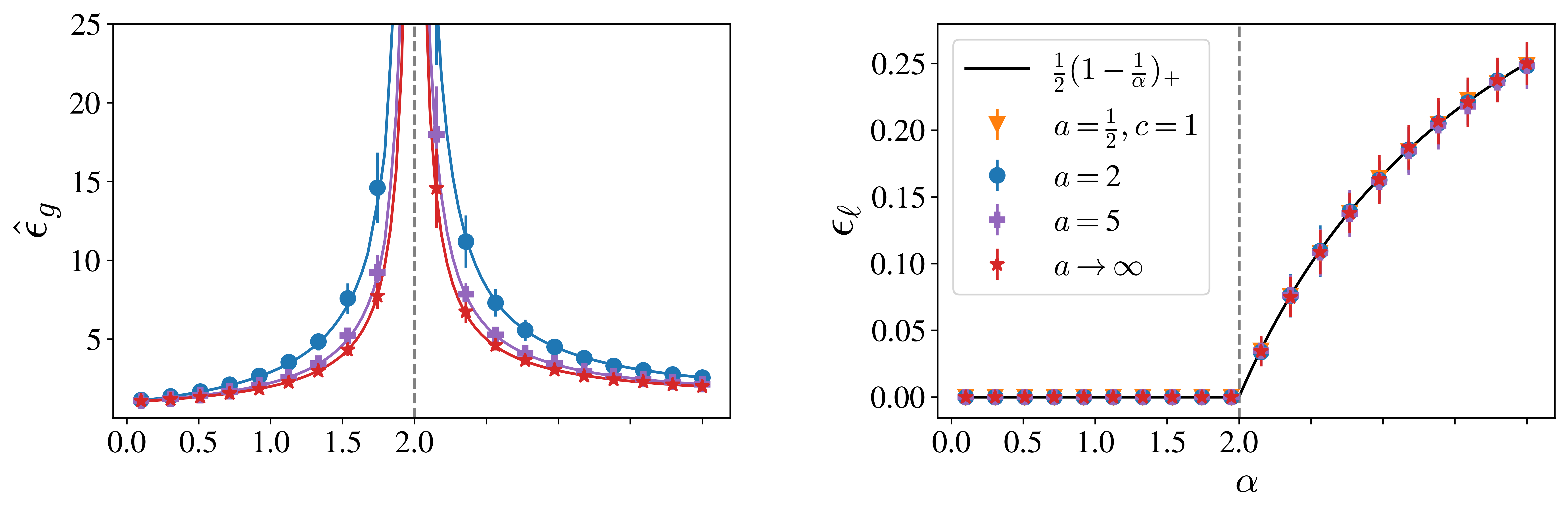}
    \caption{Test loss $\hat\epsilon_g$ (\textbf{left}) and training loss $\epsilon_\ell$ (\textbf{right}) obtained by running numerical experiments for a classification task on \textit{two} clouds with opposite means and randomly labeled points. Each cloud is generated with distribution in Eq.~\eqref{eq:invgamma_p} for different values of $a$ (i.e., different power-law decay). All clouds with $a>1$ have the same covariance $\bSigma=\bI_d$ and are obtained using Eq.~\eqref{eq:invgamma0}. The case $a=\sfrac{1}{2}$ and $c=1$, instead, corresponds to infinite $\sigma^2$ (note that in this case $\hat\epsilon_g$ is infinite and is therefore not plotted). For each $a$, the numerical experiments are compared with the theoretical prediction for a random label classification task on a single cloud centered in the origin and with the same parameter $a$, showing an excellent agreement. Note that the training loss is found to be universal and following \eqref{app:loss:gera}, independently from the variance distribution.}
    \label{app:fig:rl}
\end{figure}

\section{Note on the numerical results}\label{app:numerics}

\paragraph{State evolution equations}
Each average appearing in the update of the order parameters was performed using $10^5$ instances of the variance. For the convergence criterion, we use a tolerance of $10^{-5}$; the algorithm stops and returns the parameters after at most $10^3$ updates. 

Note that, in the case of logistic regression, there are numerical instabilities when computing $\hat v$. To avoid them we first notice that $f_k + \partial_\eta \ell_k(\Delta v f_k + \omega_k) = 0 $ holds, and we use this to compute $\partial_\omega f_k$ which we use in the application of Stein's lemma, rewriting the equation as
\begin{align}
    \hat v&=\displaystyle-\frac{\alpha\mathbb E_{k,\Delta,\zeta}\left[\sqrt\Delta f_k\zeta\right]}{\sqrt{q}} = \displaystyle-\frac{\alpha\mathbb E_{k,\Delta,\zeta}\left[\sqrt\Delta \partial_\zeta f_k \right]}{\sqrt{q}} = \alpha \mathbb E_{k,\Delta,\zeta}\left[ \frac{\Delta \partial^2_\eta\ell_k( \Delta v f_k + \omega_k)}{1+v\Delta  \partial^2_\eta\ell_k(\Delta v f_k + \omega_k)} \right],
\end{align}
where $\partial^2_\eta\ell_\pm(\eta)=\big[4 (\cosh{\eta/2})^2\big]^{-1}$ is the second derivative of the logistic loss. This equation is found to be much more stable.
\paragraph{Numerical experiments} Numerical experiments regarding the quadratic loss with ridge regularisation were performed by computing the Moore-Penrose pseudoinverse solution. For the logistic loss, we used the \texttt{LogisticRegression} module from the \texttt{Scikit-learn} package \citep{scikit-learn}. 
\end{document}